\documentclass[10pt]{article} 
\usepackage[preprint]{tmlr}

\usepackage{amsmath,amsfonts,bm}

\def\eqref#1{equation~\ref{#1}}

\def\1{\bm{1}}

\DeclareMathAlphabet{\mathsfit}{\encodingdefault}{\sfdefault}{m}{sl}
\SetMathAlphabet{\mathsfit}{bold}{\encodingdefault}{\sfdefault}{bx}{n}

\usepackage{hyperref}
\usepackage{url}

\usepackage{booktabs}
\usepackage{makecell} 
\usepackage{color, colortbl}
\definecolor{Gray}{gray}{0.9}
\newcolumntype{g}{>{\columncolor{Gray}}c}

\usepackage{xcolor}
\usepackage[linesnumbered,ruled,vlined]{algorithm2e}

\SetCommentSty{mycommfont}

\SetKwInput{KwInput}{Input}                
\SetKwInput{KwOutput}{Output}              

\usepackage{array}
\usepackage{graphicx}
\usepackage{multirow}
\usepackage[]{bm}
\usepackage{rotating}
\usepackage{subcaption}
\usepackage{wrapfig}

\title{Towards Bridging the gap between Empirical and Certified Robustness against Adversarial Examples}

\author{\name Jay Nandy \email jaynandy@comp.nus.edu.sg \\
      \addr National University of Singapore
      \AND
      \name Sudipan Saha \email sudipan.saha@tum.de \\
      \addr Technical University of Munich
      \AND
      \name Wynne Hsu\\
      \addr National University of Singapore
 	  \AND
      \name Mong Li Lee\\
      \addr National University of Singapore
	  \AND
	  \name Xiao Xiang Zhu \\
      \addr Technical University of Munich
}

\def\month{MM}  
\def\year{YYYY} 
\def\openreview{\url{https://openreview.net/forum?id=XXXX}} 

\begin{document}

\maketitle

\begin{abstract}
The current state-of-the-art defense methods against adversarial examples typically focus on improving either empirical or certified robustness. Among them, adversarially trained (AT) models produce empirical state-of-the-art defense against adversarial examples without providing any robustness guarantees for large classifiers or higher-dimensional inputs. In contrast, existing randomized smoothing based models achieve state-of-the-art certified robustness while significantly degrading the empirical robustness against adversarial examples.
In this paper, we propose a novel method, called \emph{Certification through Adaptation}, that transforms an AT model into a randomized smoothing classifier during inference to provide certified robustness for $\ell_2$ norm without affecting their empirical robustness against adversarial attacks. We also propose \emph{Auto-Noise} technique that efficiently approximates the appropriate noise levels to flexibly certify the test examples using randomized smoothing technique. Our proposed \emph{Certification through Adaptation} with \emph{Auto-Noise} technique achieves an \textit{average certified radius (ACR) scores} up to $1.102$ and $1.148$ respectively for CIFAR-10 and ImageNet datasets using AT models without affecting their empirical robustness or benign accuracy. Therefore, our paper is a step towards bridging the gap between the empirical and certified robustness against adversarial examples by achieving both using the same classifier.
\end{abstract}

\section{Introduction}

Deep neural network (DNN) based models are found to be brittle to minor, adversarially-chosen perturbations for their inputs that remain undetectable to human eyes.
A DNN classifier that correctly classifies a clean image $x$, can be easily fooled by choosing such \textit{adversarial attacks} to misclassify $x+\delta$ \citep{advStart_iclr_2014,fgsm_iclr_2015,madry_iclr_2018}. 
Here, $\delta$ is a minor \textit{adversarial perturbation} such that the change between $x$ and $x+\delta$ remains imperceptible.

Among the existing successful defense models, \textit{adversarial training} (AT) produces the best empirical robustness against the known adversarial attacks, however, without providing any guarantee \cite{madry_iclr_2018,multiple_advTrain_2019,treadAdv_icml_2019,advOverfitting_icml_2020,advTread2Gowal_2020}.
It trains a DNN classifier using strong adversaries from a specific class of perturbation (e.g., a small $\ell_p$-norm) to provide robustness for the same perturbation types.
Several certification techniques are proposed that can be applied to adversarially trained models to certifiably verify if the prediction of a test example, $x$ remains constant within its neighborhood \cite{certify1_icml_2018,certify3_nips_2018,certify9_nips_2019,certify10_uai_2018,certify17_sp_2018,certifyAbstain_iclr_2021}.
However, these certification techniques typically do not scale for larger networks (e.g., ResNet50) and datasets (e.g., {ImageNet}).
Hence, we cannot guarantee for large networks or data-sets that a powerful, not yet known attack would not break these defenses.
In fact, several recently proposed empirical defense models are later broken by stronger \textit{adaptive} adversarial attacks, indicating the importance of investigating certified defenses with suitable robustness guarantees \cite{cwAttack_sp_2017,obfuscated_icml_2018}.
In contrast to these models, the \textit{randomized smoothing} based models can provide scalable $\ell_2$-certification framework for any classification model, which is robust against large isotropic Gaussian noise \citep{certiSmoothing_icml_2019,advSmooth_nips_2019}.
However, the existing randomized smoothing-based models significantly degrade the empirical robustness compared to the state-of-the-art AT models.
In summary, a high empirical robustness along with certification guarantees are necessary to improve the reliability of a DNN model for sensitive real-world applications.
However, to the best of our knowledge, none of the existing techniques provide both high performance for both empirical robustness with such certified guarantees using the same DNN classifier.
Towards this, we aim to bridge this gap by providing robustness certification for AT models without degrading their state-of-the-art empirical robustness against adversarial examples.

\begin{table*}[t]
\centering
\resizebox{16.5cm}{!} 
{%
\begin{tabular}{l|ccccccccccccc|c}
\Xhline{3\arrayrulewidth}
\multicolumn{14}{c}{{CIFAR-10} models with the best hyper-parameters for $\ell_2$ certifications}
\\ \hline 
$\ell_2$ Radius 
& 0.0 & 0.25 & 0.5 & 0.75 & 1.0 & 1.25 & 1.5 & 1.75 & 2.0 & 2.25 & 2.5 & 2.75 & 3.0 & ACR
\\ \hline 
Baseline 
& 10.49 & 6.96 & 2.04 & 0.09 & 0.0 & 0.0 & 0.0 & 0.0 & 0.0 & 0.0 & 0.0 & 0.0 & 0.0 & 0.035
\\
\rowcolor{Gray}
Baseline + Auto-Noise 
& 33.57 & 18.56 & 10.25 & 4.44 & 0.83 & 0.07 & 0.01 & 0.0 & 0.0 & 0.0 & 0.0 & 0.0 & 0.0 & 0.124
\\
\rowcolor{Gray}
Baseline + Adaptation + Auto-Noise
& 59.64 & 21.66 & 7.81 & 3.97 & 1.28 & 0.36 & 0.07 & 0.0 & 0.0 & 0.0 & 0.0 & 0.0 & 0.0 & 0.154
\\ \hline
{Rand$_{\sigma=0.5}$ \citep{certiSmoothing_icml_2019}} 
& 62.13 & 51.68 & 40.38 & 30.25 & 20.81 & 13.36 & 7.71 & 3.38 & 0.0 & 0.0 & 0.0 & 0.0 & 0.0 & 0.494
\\ 
\rowcolor{Gray}
{Rand$_{\sigma=0.5}$ + Auto-Noise} 
& \textbf{79.48} & \textbf{71.75} & \textbf{60.23} & 48.72 & 35.97 & 25.16 & 15.09 & 10.13 & 6.98 & 5.58 & 4.18 & 2.94 & 1.76 & 0.821
\\ \rowcolor{Gray}
{Rand$_{\sigma=0.5}$ + Adaptation + Auto-Noise}
& 78.27 & 69.51 & 57.16 & 44.73 & 31.19 & 19.27 & 10.4 & 4.25 & 1.65 & 0.57 & 0.14 & 0.01 & 0.0 & 0.695
\\ \hline
{SmoothAdv$_{\sigma=0.5}$ \citep{advSmooth_nips_2019}} 
& 57.59 & 52.82 & 47.67 & 42.68 & 37.55 & 32.64 & 27.52 & 22.42 & 0.0 & 0.0 & 0.0 & 0.0 & 0.0 & 0.733\\
\rowcolor{Gray}
{SmoothAdv$_{\sigma=0.5}$ + Auto-Noise}
& 61.27 & 57.27 & 52.52 & 48.17 & \textbf{43.49} & 38.02 & 33.15 & 27.47 & 21.86 & 15.81 & 9.5 & 4.97 & 2.01 & 0.965
\\
\rowcolor{Gray}
{SmoothAdv$_{\sigma=0.5}$ + Adaptation + Auto-Noise}
& 61.23 & 56.9 & 51.33 & 46.44 & 41.05 & 35.65 & 30.11 & 24.35 & 18.48 & 12.8 & 7.76 & 4.27 & 2.3 & 0.908
\\ \hline
Adv$_{\infty}$ \citep{advOverfitting_icml_2020} 
& 13.82 & 12.22 & 10.48 & 9.12 & 7.69 & 6.32 & 5.1 & 3.79 & 0.0 & 0.0 & 0.0 & 0.0 & 0.0 & 0.154
\\
\rowcolor{Gray}
{Adv$_{\infty}$ + Auto-Noise} 
& 69.46 & 63.12 & 35.73 & 30.63 & 17.54 & 14.78 & 10.27 & 9.16 & 8.01 & 7.2 & 6.21 & 5.31 & 3.78 & 0.649
\\
\rowcolor{Gray}
{Adv$_{\infty}$ + Adaptation + Auto-Noise} 
& 70.75 & 64.54 & 50.63 & 43.38 & 32.5 & 24.43 & 18.05 & 12.2 & 8.31 & 5.37 & 3.36 & 1.96 & 1.23 & 0.76
\\ \hline
Adv$_2$ \citep{advOverfitting_icml_2020}
& 30.37 & 26.98 & 23.98 & 21.35 & 18.4 & 15.94 & 13.52 & 10.63 & 0.0 & 0.0 & 0.0 & 0.0 & 0.0 & 0.367
\\
\rowcolor{Gray}
{Adv$_2$ + Auto-Noise} 
& 64.45 & 60.57 & 45.73 & 41.06 & 28.48 & 22.92 & 15.1 & 10.77 & 7.23 & 4.8 & 2.77 & 1.67 & 1.1 & 0.702
\\
\rowcolor{Gray}
{Adv$_2$ + Adaptation + Auto-Noise} 
& 61.96 & 58.58 & 53.64 & \textbf{49.67} & 42.76 & \textbf{38.69} & \textbf{34.54} & \textbf{30.36} & \textbf{24.65} & \textbf{20.77} & \textbf{17.09} & \textbf{13.66} & \textbf{9.18} & \textbf{1.102}
\\ \hline
Marcer$_{\sigma=0.5}$ \citep{macer_iclr_2020} 
& 64.2 & 57.5 & 49.9 & 42.3 & 34.8 & 27.6 & 20.2 & 12.6 & 0.0 & 0.0 & 0.0 & 0.0 & 0.0 & 0.691 \\

Consistency$_{\sigma=0.5}$ \citep{consistency_nips_2020} 
& 52.3 & 48.9 & 45.1 & 41.3 & 37.8 & 33.9 & 29.9 & 25.2 & 0.0 & 0.0 & 0.0 & 0.0 & 0.0 & 0.726\\

Boosting$_{\sigma=0.5}$ \citep{boostingRS_iclr_2022}
& 65.0 & 59.0 & 49.4 & 44.8 & 38.6 & 32.0 & 26.2 & 19.8 & 0.0 & 0.0 & 0.0 & 0.0 & 0.0 & 0.756\\ 
\Xhline{3\arrayrulewidth}
\end{tabular}%
}
\vspace*{-0.5em}
\caption{ {CIFAR-10}: Certified accuracy at different $\ell_2$ radii and ACR scores using the best hyper-parameters.
Existing randomized-smoothing based models are trained and certified using noise-level, $\sigma=0.5$.
We also present the best reported results for Marcer and Consistency at \citep{consistency_nips_2020} and for Boosting at \citep{boostingRS_iclr_2022} for our comparisons.
For detailed results and hyper-parameters for both {ImageNet} and {CIFAR-10} datasets, please refer to Table \ref{table_app_imagenet_certify} and \ref{table_app_cifar10_certify} respectively in Appendix.
}
\label{table_cifar10_certify}
\end{table*}

In this paper, we propose a novel \textit{certification through adaptation} framework that transforms an AT model into a randomized smoothing framework during inference to provide non-trivial $\ell_2$ certification without any additional training or architecture modifications.
Our proposed certification technique consists of two steps:
We first adapt the AT model using popular batch normalization adaptation technique using appropriate levels of Gaussian noises separately for each test example  \citep{cariucci2017autodial,li2016revisiting}.
This process significantly boosts the performance of the AT models against the random isotropic Gaussian noises.
Hence, we can now directly apply the \textit{randomized smoothing} based certification technique to provide $\ell_2$ certification in the next step.
However, choosing the Gaussian noise for each test example is a challenging task.
The existing randomized smoothing based models that use Gaussian noises for training, use the same noise levels to certify each test example, significantly compromising their certification performance.
Towards this, we also propose an \textit{Auto-Noise} technique to efficiently approximate the appropriate Gaussian noise levels for correctly certifying each test example during inference.

In the following, we summarize the list of contributions for our paper:
\begin{enumerate}
\item We propose a novel \textit{certification through adaptation} framework that adapts an AT model using Gaussian noises to provide non-trivial robustness certifications at large $\ell_2$ radii.
Our proposed technique only requires a set of \textit{clean images}, obtained from training/validation or test set to adapt the AT models using popular BN-adaptation technique \citep{cariucci2017autodial,li2016revisiting}.

\item We also propose \textit{Auto-Noise technique} to efficiently approximate the appropriate Gaussian noise levels for certifying each test example during inference.
Auto-Noise is applicable even for existing randomized smoothing based models and often significantly improves the certification performance.
Our \emph{Certification through Adaptation} together with \emph{Auto-Noise} technique produces \textit{average certified radius (ACR) scores} upto $1.102$ and $1.148$ for CIFAR-10 and ImageNet for AT models, achieving the state-of-the-art performance for CIFAR-10.
Notably, our proposed method is applied during inference, without affecting the empirical robustness or benign accuracy of AT models to produce these non-trivial $\ell_2$ certification results.

\item Our results also suggests a stronger correlation between empirical and certified robustness that empirically stronger AT models also produce better $\ell_2$ certification performance. 
\end{enumerate}

\section{Related Work}
\subsection{Adversarial Robustness for DNN models}
\textbf{Empirical Defenses and Adversarial Training. }
Defense models against adversarial examples can be broadly categorized as: \textit{empirical} and \textit{certified} defenses.
Empirical defenses demonstrate empirical robustness against adversarial attacks, typically without out providing any certification guarantees  \citep{mnist_multiple_iclr2019,cure_cvpr_2019,rbfcnn_ijcnn_2020,reversible_iccv_2021}.
\textit{Adversarial training} achieves the state-of-the-art empirical defense \citep{madry_iclr_2018}.
It optimizes the following loss function for a DNN classifier, $f$, to provide robustness within an $\epsilon$-bounded \textit{threat model} for an $\ell_p$ norm, where the perturbations, $\delta \in \Delta$ are constrained as $\Delta = \{ \delta : ||\delta||_p \leq \epsilon\}$: 
\begin{equation}
\label{eq_advTrain}
\min_{\theta} \mathbb{E}_{(x,y)}[\max_{\delta \in \Delta} \mathcal{L}(f_{\theta}(x+\delta), y)]
\end{equation}
where, $\theta$ denotes the model parameters. $\mathcal{L}$ is the classification loss.

The \textit{inner maximization} in Eq. \ref{eq_advTrain} is solved by producing adversarial examples using strong iterative adversaries, e.g., \textit{projected gradient descent (PGD)} attack \citep{advTrain_arxiv_2017,madry_iclr_2018}.
\cite{fgsmAdvTrain_iclr_2020} found that even a single-step \textit{fast gradient sign method (FGSM)} attack-based AT models also achieves high empirical robustness \citep{fgsm_iclr_2015}.
\cite{fat_icml_2020} proposed to use the least adversaries for training.
Recently TRADES \citep{treadAdv_icml_2019}, Adv-LLR \citep{llrAdv_nips_2019} introduced additional regularizers to achieve higher empirical robustness by smoothing the loss surface.
\cite{advOverfitting_icml_2020} showed that even the standard PGD based AT model with early-stopping criteria provides one of the best empirical defenses for a given perturbation type.
Recent works also explored the importance of different hyper-parameters for adversarial training \citep{advTread2Gowal_2020,advTricks_iclr_2021} as well as incorporating additional data in a semi-supervised fashion \citep{advUnlabelled1_arxiv_2019,advUnlabelled2_arxiv_2019} to further improve their empirical robustness against adversarial attacks.
Recently, \cite{advTrain_cprobust_arxiv21} also demonstrated that adversarial training with smaller perturbation can also improve the performance against random corruptions.

\smallskip
\textbf{Certified Defenses. }
Empirical defenses demonstrate robustness only against the \textit{known} adversaries without providing any guarantees.
In fact, most empirical defenses proposed in the literature were later \textit{broken} by stronger adversaries, highlighting the importance of certified defenses to provide robustness guarantees \citep{obfuscated_icml_2018,spsa_icml_2018,overpoweredAttack_2017}.

Several recent works proposed to train neural network models with provable robustness guarantees. 
These works include methods based on semidefinite relaxations \citep{certify4_iclr_2018}, linear relaxations and duality \citep{certify1_icml_2018,certify8_nips_2018}, abstract interpretation \citep{certify6_icml_2018}, and interval bound propagation \citep{certify16_arxiv_2018}.
Parallel to training a certified defense, several works also focus on certifying the already trained models \citep{exact1_iclr_2017,certify17_sp_2018,certify14_icml_2018,certify3_nips_2018,bunel_nips_2018}.
Recently \cite{certifyCombiner_iclr_2021} combined a \textit{small} certification network with a \textit{large}, empirically robust AT model using some selection criteria to boost overall benign accuracy along with empirical robustness for the certified framework.
However, most of these techniques do not scale well for large DNN classifiers (e.g., ResNet50) or higher-dimensional datasets (e.g., {ImageNet}).

\smallskip
\textbf{Randomized Smoothing} is a promising certification technique that can be scaled to larger networks and higher-dimensional datasets.
It was initially proposed as a heuristic defense \citep{smooth1_heu_2017,smooth2_heu_2018} and later shown to be certifiable \citep{randSmoothDP_sp_2019,certiSmoothing_nips_2019}.
Recently, \cite{certiSmoothing_icml_2019} and \cite{advSmooth_nips_2019} separately provided strong robustness guarantees for  $\ell_2$-norm. 
A randomized smoothing based certification model requires their base-classifier to be robust against large Gaussian perturbations to produce non-trivial results.
\cite{certiSmoothing_icml_2019} proposed to train their base-classifier by incorporating random Gaussian noises.
Several recent works focused on improving the base classifiers to achieve better certification performance by adversarially choosing the noise \citep{advSmooth_nips_2019}, incorporating additional regularizers \citep{macer_iclr_2020,consistency_nips_2020}, by ensembling multiple base-models \citep{boostingRS_iclr_2022} etc.
Several works also investigated on improving certification guarantees using different smoothing measures \citep{certiSmoothing_nips_2019,certiLP_arxiv_2019,anyLpCertify_arxiv_2020} or divergences \citep{certiLP_iclr_2020}.
\cite{advDenoised_nips_2020,carlini2022certified} demonstrated that we can achieve non-trivial certified robustness even for a standard DNN classifier by incorporating an additional denoising module as a pre-processing unit.
Notably, randomized smoothing is the only scalable certification framework and also provides superior performance for different perturbation types \citep{certiLP_iclr_2020}.

However, while achieving the state-of-the-art certification performance, randomized smoothing significantly degrades the empirical robustness against adversarial attacks compared to the state-of-the-art AT models \citep{randSmoothDP_sp_2019,advSmooth_nips_2019,certiSmoothing_icml_2019}.
Towards this, our proposed technique transforms an AT model into a randomized smoothing classifier without requiring additional training or architectural modification.
Since AT models already provide the state-of-the-art empirical defense, we achieve both empirical and certified robustness against adversarial examples using the same classifier.

\smallskip
\textbf{Batch-normalization and Robustness. }
Several recent papers investigate the effects of batch-normalization layers for different aspects of robustness.
Many of these works focused on improving robustness against random corruptions by adapting batch-normalization using a sufficiently large set of test images from the same covariate shift \citep{adaptBN_cp_nips_2020,adaptBN_cp_arxiv_2020,adaptBN_cp_wacv_2021}.
By hypothesizing that clean and adversarial examples belongs to different domains, several recent works proposed to apply different branches of BN to separately capture their distributions \citep{auxBN_cvpr_2020,mixtureBN_iclr2020,dualBN_1,dualBN_2,dualBN_3}.
%
\cite{bnreduce_advrobust_iccv21} presents empirical evidence to argue that BN shifts a model towards being more dependent on non-robust features (NRFs).
Unlike these previous works, we proposed to adapt BN layers using appropriate Gaussian noise levels to provide $\ell_2$ certified robustness for AT models.

\subsection{Test-time Adaptation \& applications}
Test-time adaptation techniques have been widely explored before in the field of domain adaptation \citep{da1_2017,roy2019unsupervised,huang2018decorrelated,li2016revisiting} and covariate-shift adaptation \citep{adapt_cp_icml_2020,adaptation2_2017,adaptation3_cvpr_2020,adaptation4_2020,adaptBN_cp_nips_2020,adaptBN_cp_arxiv_2020,adaptBN_cp_wacv_2021}.
However, to the best of our knowledge, such techniques are never applied for adversarial robustness and certification.
Our paper mainly focuses on one of the most popular and effective mechanisms, called \textit{adaptive batch-normalization}.

A batch-normalization (BN) layer computes the mean and variance of the hidden activation maps across the channels to normalize these activations to $\mathcal{N}(0,1)$ before feeding into the next hidden layer \citep{bn_icml_2015}.
This process reduces the dependencies among different hidden layers, improving the training efficiency for deep architectures.
However, the distributional shifts in the test examples lead to different activation statistics compared to the training examples.
Hence, the statistics estimated during training fail to correctly normalize the activation tensors to $\mathcal{N}(0,1)$. 
Consequently, it breaks the crucial assumption for the subsequent hidden layers to work.
Adaptive BN technique computes the BN statistics from the feature activations, $\mu_t$, $s_t^2$, of the test batch.
We can adapt them with the existing \textit{training} statistics, $\mu_T$, $s_T^2$, learned using the training batches as \citep{cariucci2017autodial,li2016revisiting,adaptBN_cp_nips_2020}: 
\begin{equation}
\label{eq_adaptStat}
\overline{\mu} = \rho \cdot \mu_t + (1-\rho)\cdot \mu_T \quad
\overline{s} = \rho \cdot s_t + (1-\rho) \cdot s_T
\end{equation} 

where, $\rho \in [0,1]$ is the momentum.
The choice of $\rho=0$ is equivalent to the standard inference setup with a deterministic DNN classifier in the IID settings.
We should choose $\rho=1$  when receiving larger test batches as it can provide a better estimation of the test distributions.

\smallskip
\textbf{Assumptions and Limitations. }
The existing BN-adaptation techniques typically require a \textit{large set of test images} from the same ``unknown” test distributions.
However, this assumption may not hold for several real-world applications, e.g., stateless web APIs.
Also, these test images should be \textit{semantically diverse}, preferably over multiple classes, to effectively estimate the test distributions.
Hence, it further limits the practical usability of these frameworks for real-world applications, e.g., autonomous cars.

Unlike these models for domain adaptation and corruption robustness, our proposed certification framework does not make any such assumptions. 
As we shall see that we can appropriately approximate the required Gaussian noise level for adaptation to certify a test image.
Therefore, we can pre-select a diverse set of clean images, $\mathbf{X}_{clean}$ and inject the random Gaussian noises to appropriately adapt the models as required, addressing both of these limitations.

\section{Proposed Methodology}
In this section, we first present the background of the randomized smoothing technique and explain why it is not directly applicable to AT models.
Next, we present our proposed \textit{certification through adaptation} framework that adapts a DNN model during inference to provide certified robustness without additional training or architectural modifications.

\subsection{Background on Randomized Smoothing}
\label{sec_rand_smooth}
Consider a classification model, $f$, that maps inputs in $\mathbb{R}^d$ to $\mathcal{Y}$ classes.
The randomized smoothing framework transforms the original base classifier, $f$ into a new, smoothed classifier $g$ \cite{certiSmoothing_icml_2019}.
In particular, for an input $x \in \mathbb{R}^d$, the smoothed classifier $g$ returns the most probable class to be predicted by the base classifier $f$ under isotropic Gaussian noises of $x$. That is,
\begin{equation}
\label{eq_smoothing}
g(x) = arg\max_{y\in \mathcal{Y}} \mathbb{P}(f(x+\delta)==y) \quad \text{s.t.} ~~\delta \sim \mathcal{N}(0, \sigma^2I). 
\end{equation}
where, $\sigma^2I$ is the covariance matrix and $\sigma$ denotes the noise level for certifying $x$.
$\sigma$ controls the trade-off between robustness at different $\ell_2$ radii: Increasing $\sigma$ improves the robustness of $g$ at higher $\ell_2$ radii. 
However, it degrades the robustness at smaller $\ell_2$ radii.

\cite{certiSmoothing_icml_2019} presented a tight robustness guarantee using Neyman-Pearson lemma for the smoothed classifier, $g$ and provided an efficient algorithm using Monte Carlo sampling for certification. 
We can also obtain the same guarantee by explicitly computing the Lipschitz constant of the smoothed classifier as shown in \citep{advSmooth_nips_2019,anyLpCertify_arxiv_2020}.
The certification procedure is as follows: 
Suppose a base classifier $f$ classifies $\mathcal{N}(x, \sigma^2I)$ to return the \textit{``most probable"} class, $c_A$ with probability $p_A = \mathbb{P}(f(x+\delta)==c_A)$ and the \textit{``runner-up"} class $c_B$ with probability $p_B = \max_{y\neq c_A} \mathbb{P}(f(x+\delta)==y)$.
Then, the smooth classifier, $g$ is certifiably robust around $x$ within an $\ell_2$ radius of $R$, as follows:
\begin{equation}
\label{eq_certify_l2}
R = \frac{\sigma}{2}\Big(\Phi^{-1}(p_A)-\Phi^{-1}(p_B)\Big)
\end{equation}
where, $\Phi^{-1}$ denotes the inverse of the standard Gaussian CDF.

However, computing the exact values for $p_A$ and $p_B$ is impossible in practice when $f$ is a DNN.
\cite{certiSmoothing_icml_2019} addressed this problem using Monte Carlo sampling to estimate $\underline{p_A}$ and $\overline{p_B}$ such that $\underline{p_A} \leq p_A$ and $\overline{p_B} \geq p_B$ with arbitrarily high probability.
The certified radius for input $x$ is then computed by replacing $p_A$ and $p_B$ with $\underline{p_A}$ and $\overline{p_B}$ respectively in Eq. \ref{eq_certify_l2}.

As we can see in Equation \ref{eq_smoothing} that the base classifier, $f$ needs to be robust against large Gaussian noises to produce non-trivial robustness certification results. 
Otherwise, it leads to lower $p_A$ and hence a lower certification of $R$ for the test examples.
Existing randomized smoothing-based models applies custom-trained using explicit Gaussian noises to learn their original base classifier \citep{randSmoothDP_sp_2019,certiSmoothing_icml_2019,advSmooth_nips_2019,macer_iclr_2020,consistency_nips_2020}.
However, these models produce significantly lower empirical robustness compared to the AT models \citep{madry_iclr_2018,treadAdv_icml_2019,advOverfitting_icml_2020,advTread2Gowal_2020}.
In contrast, AT models are not robust against large Gaussian noises in the standard inference settings \citep{cpGaussianAug_icml_2019}. 
In our paper, we bridge this gap between these two research directions by certifying the AT models through adaptation, as described in the following.

\begin{algorithm}[H]
\DontPrintSemicolon
  
  \KwInput{ $f$: classifier, \quad $x_{test}$: test example, \quad $\sigma$: desired noise-level, 
  \quad $\mathbf{X}_{clean}$: batch of clean images sampled from train/validation data or test stream,
  \quad $N$: No. of noisy samples for Monte-Carlo estimation (Eq. \ref{eq_certify_l2})}

  \KwOutput{ Certifiably robust $\ell_2$ radius of $R$ for $x_{test}$.}
  
  \vspace{0.5em}
  \tcc{ Step 1: Adapt BN parameters using $\mathbf{X}_{clean}$ with $\rho=1$ (Eqn \ref{eq_adaptStat}).}

	$\tilde{\mathbf{X}}_{clean} = [x + \mathcal{N}(0, \sigma I)~~\forall~x~\in~\mathbf{X}_{clean}]$
	\tcp*{perturb $\mathbf{X}_{clean}$ with random noise.}	

  	$f_{adapt}$ = \textsc{Clone}$(f.train())$ 
	\tcp*{clone $f$ with \textit{train-mode}.}

	$\ f_{adapt}(\tilde{\mathbf{X}}_{clean})$ 
	\tcp*{forward pass for BN parameter adaptation.}
	
	$f_{adapt}.eval()$ 
	\tcp*{fix the parameters.}

  \vspace{0.5em}
  \tcc{ Step 2: Certify $x_{test}$ using Randomized Smoothing framework.}
	$g$ = \textsc{getRandomizedModel($f_{adapt}$)}
	\tcp*{Convert $f_{adapt}$ to randomized-smoothing classifier $g$	(Eqn \ref{eq_smoothing}).}

	$R$ = \textsc{Certify}($g$, $x_{test}$; $\sigma$, $N$)
	\tcp*{Execute \ref{eq_certify_l2} for $\ell_2$ certification.}

	return $R$
	\tcp*{Return $\ell_2$ certification for $x_{test}$.}

\caption{\textsc{Certification-through-Adaptation}}
\label{algo_certify}
\end{algorithm}

\subsection{Proposed Certification through Adaptation}
Our proposed \textit{certification through adaptation} framework consists of two steps:
Given a test image $x_{test}$, we first adapt the original classification model, $f$ using adaptive BN technique with an appropriate/ pre-selected level of Gaussian noise, $\sigma$ to obtain the base classifier, $f_{adapt}$ for certification using randomized smoothing.
Next, we freeze the model parameters and use the fixed adapted model, $f_{adapt}$, as our base classifier to certify the test example, $x_{test}$ using randomized smoothing technique (Equation \ref{eq_certify_l2}).
The proposed \textit{certification through adaptation} method is presented in Algorithm \ref{algo_certify}. 

\smallskip
\begin{wrapfigure}{r}{7.0cm}
\centering
\includegraphics[width=6.3cm, height=4.5cm]{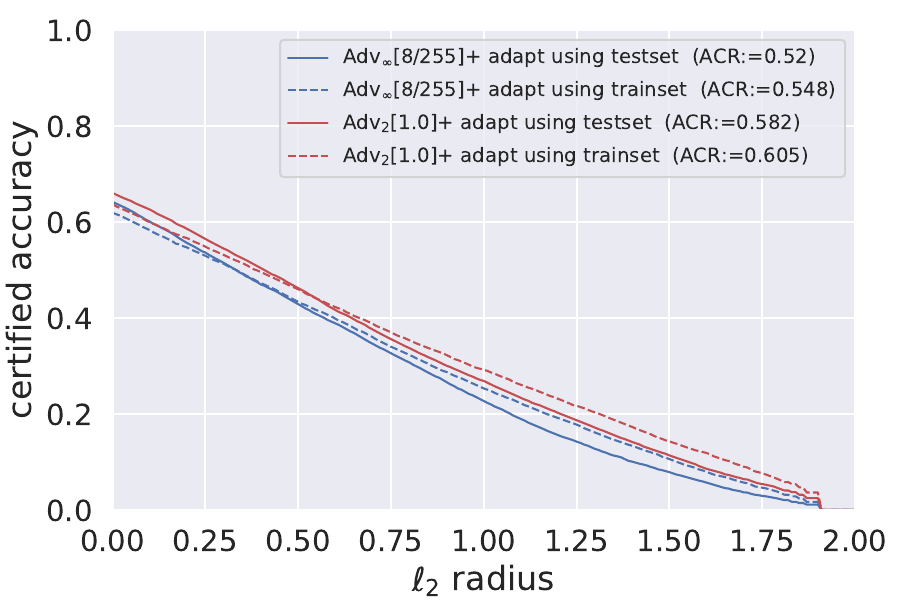}
\vspace{-0.5em}
\caption{ Certification performance on CIFAR-10 as we apply training data vs test data for adaptation in Algorithm \ref{algo_certify}.
Adv$_{\infty}[8/255]$ and Adv$_2[1.0]$ respectively denote AT models, trained at $\ell_{\infty} \leq 8/255$ and $\ell_2 \leq 1.0$ threat-boundaries.
}
\label{fig:train_test}
\end{wrapfigure}
\textbf{Training versus Test images for model adaptation.}
Recall that adaptive BN requires a large set of diverse test images to correctly re-estimate the BN layer statistics. 
However, to provide certification for $\ell_2$-norm, we only need to adapt our model against Gaussian perturbations. 
Hence, unlike existing test-time adaptation-based models for covariate shift or domain adaptation problems, we can pre-select a sufficiently large and diverse set of clean images, $\mathbf{X}_{clean}$. 
We apply the selected level of Gaussian noise, $\sigma$ to obtain the base classifier, $f_{adapt}$ for certifying test examples, $x_{test}$.
Notably, we can sample $\mathbf{X}_{clean}$ from training/validation or test sets. 
Since the underlying distribution of these clean images remains the same, it does not affect the certification performance.
In Figure \ref{fig:train_test}, we compare the performance as we randomly sample $\mathbf{X}_{clean}$ from training vs. test sets for CIFAR-10.
We can see that the certification performance of AT models remains almost the same in both cases.
The slight differences in their performances typically arise due to the underlying randomization of the set, $\mathbf{X}_{clean}$ and added Gaussian noise to estimate the BN parameters.

\smallskip
\textbf{Effect on empirical robustness and benign (clean) accuracy. }
The main advantage of our proposed framework is that we can provide certification for AT models without affecting their state-of-the-art empirical robustness and benign accuracy.
Given a test example, we first obtain the predicted class label from the original classifier, $f$ (i.e., without making any change). 
Next, we adapt the model to $f_{adapt}$ using appropriate $\sigma$ to certify the classification prediction using proposed Algorithm \ref{algo_certify}.
Hence, we maintain the same empirical robustness and benign accuracy as reported in the  existing papers \citep{advOverfitting_icml_2020,madry_iclr_2018}.

\smallskip
\textbf{Applicability. }
Our proposed ``certification through adaptation" technique can be applied to any classification model, $f$ with batch-normalization layers.
However, achieving high accuracy against large random Gaussian perturbations is a necessary condition: a randomized smoothing classifier, $g$  needs to \textit{consistently} predict the correct class to provide higher certification guarantees at larger radii.
Hence for standard non-robust DNN classifiers, we can only achieve higher $\ell_2$ certification guarantees at very small $\ell_2$ radii  (see in Table \ref{table_app_imagenet_certify} and Table \ref{table_app_cifar10_certify}).
Further,we also observe that adapting the existing randomized smoothing models does not necessarily improve the overall certification performance (see Table \ref{table_cifar10_certify} and Figure \ref{fig:certify_autoNoise_rs} in Section \ref{sec_autonoise_experiment})
In contrast, AT models with our proposed ``offline" adaptation significantly improve their performance against large Gaussian perturbations, providing non-trivial certification robustness.

\subsection{Proposed Auto-Noise: Appropriate Noise level for Certification}  
Robustness of a classification model can significantly vary at different input spaces.
Hence, choosing appropriate noise-level for certifying is an important but challenging task for randomized smoothing based certification techniques. 
While choosing a lower noise level produces significantly lower-estimates of certified radii, over-estimation of noise may fail to provide any certification robustness for a test example.
A brute-force approach to address this problem would be to evaluate the certification results on multiple noise levels and report the maximum certified $\ell_2$ radii.
However, certification using randomized smoothing is an extremely time-consuming process: it requires to evaluate a large number of noisy samples (of cardinality $N=100,000$) to estimate $\underline{p_A}$ and $\overline{p_B}$ using Monte-Carlo sampling (Eq. \ref{eq_certify_l2}).
For example, it can take upto $110$ seconds to certify an ImageNet test example with ResNet-50 models on NVIDIA RTX 2080 Ti \citep{certiSmoothing_icml_2019}.
Notably, as we choose a smaller set of noisy samples to estimate $\underline{p_A}$ and $\overline{p_B}$, it provides significantly lower $\ell_2$ certified radii.
Hence, existing randomized smoothing based models typically use the same Gaussian noise level as applied to train their base classifiers \citep{certiSmoothing_icml_2019,advSmooth_nips_2019,macer_iclr_2020,consistency_nips_2020}. 
However, we demonstrate that it significantly underestimates the certification performance of the randomized smoothing framework.
Here, we present a simple but effective Auto-Noise technique to choose appropriate $\sigma$ for a given test example, $x_{test}$.

Auto-Noise technique aims to approximate the appropriate noise-level from a given set, 
${\bm \sigma} = \{\sigma_1, \sigma_2, \cdots\}$. 
The key idea of our proposed Auto-Noise method is as follows: 
\textit{Although a small set of noisy examples underestimates the $\ell_2$ certified noise levels, it can provide a fair comparison of different noise levels for certifying $x_{test}$.}
We propose to use a small set of $N_{auto}$ noisy examples, $N_{auto}$ to approximately obtain the $\ell_2$ certification radii for different noise levels, $\sigma \in \bm{\sigma}$.
We select the noise level $\sigma_{auto}$ that produces the maximum $\ell_2$ radius using $N_{auto}$ noisy examples, i.e.:
\begin{equation}
    \sigma_{auto} = arg\max_{\sigma \in {\bm \sigma}} \quad \textsc{Certify}(g, x_{test}; \sigma, N_{auto})
\end{equation}
where, $g$ denotes the randomized-smoothing classifier obtained using the base classifier, $f$ (Eq. \ref{eq_smoothing}). 
For AT models, we should adapt the models, $f_{adapt}$ with noise level $\sigma$ as the base classifier.

Finally, we use $\sigma_{auto}$ for certifying $x_{test}$ with a large number of noisy samples, $N = 100,000$. 

\smallskip
\textbf{Computational Overhead. }
In practice, the ideal choices of noise levels as ${\bm \sigma}$ remains reasonably small.
For example, in our experiments, we select ${\bm \sigma} = \{0.12, 0.25, 0.37, 0.50, 0.67, 0.75, 0.87, 1.0\}$, i.e. of cardinality=$8$ and set $N_{auto}=1,000$.
Hence, we require an additional $8,000$ iterations to obtain the appropriate $\sigma$ for each test-examples, along with $100,000$ iterations to get the final certification.
In other words, with very little computational overhead, we can approximate the appropriate noise levels for each test example. 

Note that our Auto-Noise algorithm using a small set, $N_{auto}=1,000$ may not provide reliable estimation of the most appropriate noise-level, $\sigma$.
However, as shown in Table \ref{table_cifar10_certify}, it significantly improves the certification performance for both AT models and the existing randomized smoothing based models, compared to the fixed choices of $\sigma$ \citep{certiSmoothing_icml_2019,certiSmoothing_nips_2019}.
Furthermore, by using certification through adaptation along with Auto-Noise method, we can produce state-of-the-art certification performance for AT models, trained using $\ell_2$ bounded adversarial examples.


\section{Experimental Results}
\label{sec_experiment}
\textbf{Setup. }
We use {CIFAR-10} \citep{db_cifar10} and {ImageNet} \citep{db_imagenet} datasets for our experiments.
For {CIFAR-10}, we use pre-activation ResNet18 and for {ImageNet}, we use ResNet50  \citep{resnet_cvpr_2016,resnet_eccv_2016}.
For our experiments, we train the AT models using the \textit{early stopping} criteria \citep{advOverfitting_icml_2020}.
For {ImageNet}, we use two AT models, Adv$_{\infty} [4/255]$ and Adv$_{2} [3]$, learned at $\ell_{\infty}$ and $\ell_2$ threat models  with threat boundaries of $4/255$ and $3$ respectively.
For {CIFAR-10}, we train multiple AT models with different threat boundaries.
We denote them by incorporating their corresponding threat boundaries, applied for training.
For example, we denote an AT model, trained with threat boundary of $8/255$ as Adv$_{\infty} [8/255]$.

For our comparisons, we use the standard DNN Baseline and Rand$_{\sigma=0.5}$ models.
Baseline models are trained using clean images.
Rand$_{\sigma=0.5}$ models are trained by augmenting random noise, sampled from $\mathcal{N}(0,\sigma^2I)$ with $\sigma=0.5$ \cite{certiSmoothing_icml_2019}.
We also compare with the current state-of-the-art certification models, SmoothAdv for CIFAR-10 \citep{advSmooth_nips_2019}.
Please refer to Appendix \ref{sec_implementation} for additional details
\footnote{For  {ImageNet}, we obtain Adv$_{\infty}$ and Adv$_{2}$ from \href{https://github.com/locuslab/robust_overfitting}{https://github.com/locuslab/robust\_overfitting} and
Baseline and Rand$_{\sigma=0.5}$ models from  \href{https://github.com/locuslab/smoothing}{https://github.com/locuslab/smoothing}.}.

\begin{table*}[h]
\centering
\resizebox{15.5cm}{!}{%
\begin{tabular}{l!{\vrule width 1.5pt}cccccl!{\vrule width 1.5pt}cccc}
\cmidrule[1.2pt]{1-5} \cmidrule[1.2pt]{7-11}
\multicolumn{5}{c}{(a) {ImageNet}} & $\qquad$ &\multicolumn{5}{c}{(b) {CIFAR-10}} \\
\cmidrule[1.2pt]{1-5} \cmidrule[1.2pt]{7-11}
Model 
& $\sigma=0$ & $\sigma=0.25$ & $\sigma=0.5$ & $\sigma=0.75$ &
& Model 
& $\sigma=0$ & $\sigma=0.25$ & $\sigma=0.5$ & $\sigma=0.75$ 
\\ \cline{0-4} \cline{7-11}
Baseline 
& \textbf{75.2{\tiny $\pm0.0$}} & 11.8{\tiny $\pm0.22$}& 0.3{\tiny $\pm0.01$}& 0.1{\tiny $\pm0.0$}&

& Baseline 
& \textbf{95.2{\tiny $\pm0.0$}}& 10.9{\tiny $\pm0.88$}& 10.6{\tiny $\pm0.76$}& 10.5{\tiny $\pm1.19$} \\

+ adaptive BN 
& 74.4{\tiny $\pm0.04$}& \textbf{31.0{\tiny $\pm0.27$}} & \textbf{7.7{\tiny $\pm0.24$}} & \textbf{2.4{\tiny $\pm0.01$}} &

& + adaptive BN 
& 95.0{\tiny $\pm0.57$}& \textbf{40.1{\tiny $\pm0.97$}} & \textbf{22.0{\tiny $\pm0.83$}} & \textbf{17.2{\tiny $\pm0.66$}}

\\ \cline{1-5} \cline{7-11}
Adv$_{\infty} [4/255]$
& \textbf{62.8{\tiny $\pm0.0$}}& 3.9{\tiny $\pm0.03$}& 0.4{\tiny $\pm0.01$}& 0.2{\tiny $\pm0.01$} &
& Adv$_{\infty} [8/255]$ 
& \textbf{82.1{\tiny $\pm0.0$}}& 40.2{\tiny $\pm4.56$}& 16.1{\tiny $\pm7.85$}& 12.2{\tiny $\pm5.23$} 
\\
\cellcolor{Gray}+ adaptive BN
& \cellcolor{Gray} 60.8{\tiny $\pm0.16$}& \cellcolor{Gray} \textbf{53.4{\tiny $\pm0.15$}}& 
\cellcolor{Gray} \textbf{44.9{\tiny $\pm0.08$}}& \cellcolor{Gray} \textbf{33.7{\tiny $\pm0.28$}} &
& \cellcolor{Gray} + adaptive BN
& \cellcolor{Gray} 81.6{\tiny $\pm0.96$}& \cellcolor{Gray} \textbf{74.2{\tiny $\pm0.95$}}& 
\cellcolor{Gray} \textbf{62.4{\tiny $\pm0.64$}}& \cellcolor{Gray} \textbf{51.0{\tiny $\pm1.03$}}
\\ \cline{1-5} \cline{7-11}

Adv$_{2} [3]$ 
& \textbf{59.8{\tiny $\pm0.0$}} & 9.8{\tiny $\pm0.08$}& 0.9{\tiny $\pm0.01$}& 0.3{\tiny $\pm0.0$} &
& Adv$_{2}  [1]$
& \textbf{81.6{\tiny $\pm0.0$}} & 47.5{\tiny $\pm5.1$}& 21.5{\tiny $\pm7.79$}& 14.3{\tiny $\pm5.63$} 
\\
\cellcolor{Gray} + adaptive BN
& \cellcolor{Gray} 58.3{\tiny $\pm0.08$}& \cellcolor{Gray} \textbf{53.7{\tiny $\pm0.14$}} & 
\cellcolor{Gray} \textbf{47.3{\tiny $\pm0.14$}} & \cellcolor{Gray} \textbf{39.8{\tiny $\pm0.18$}} &
& \cellcolor{Gray} + adaptive BN
& \cellcolor{Gray} 81.8{\tiny $\pm0.7$}& \cellcolor{Gray} \textbf{75.8{\tiny $\pm0.43$}} & 
\cellcolor{Gray} \textbf{64.9{\tiny $\pm0.73$}} & \cellcolor{Gray} \textbf{53.5{\tiny $\pm1.71$}} 
\\ 

\cmidrule[1.2pt]{1-5} \cmidrule[1.2pt]{7-11}
\end{tabular}%
}
\vspace{-0.5em}
\caption{ Top-1 accuracy of AT models significantly improve using adaptive BN under different levels of Gaussian noises (when $\sigma>0$).
We randomly sample the noises and report $(mean~\pm~2\times sd)$ for five different runs.
}
\label{table:gaussian}
\vspace{-0.5em}
\end{table*}

\subsection{Performance under Gaussian Noise. }
We first investigate the performance of AT models as we significantly increase the Gaussian noises.
As we note in Section \ref{sec_rand_smooth}, it is a necessary condition to provide non-trivial robustness certification at larger $\ell_2$ radii.
In Table \ref{table:gaussian}, we present a comparative performance for Baseline, Adv$_{\infty}$, and Adv$_{2}$ models for both ImageNet and CIFAR-10 datasets.
We can see that the classification performance of all these models sharply degrades under large Gaussian noises in standard inference settings.
However, we can improve these performances by adapting them under the same level of Gaussian noises using adaptive BN techniques.
In particular, we observe that AT models achieve significantly higher performance gain using adaptive BN than the non-robust, Baseline models under Gaussian noise levels.
For example, at $\sigma= 0.5$, Baseline, Adv$_2 [3]$ and Adv$_{\infty} [4/255]$ respectively achieve top-1 accuracy of $0.3\%$, $0.4\%$, and $0.9\%$  for {ImageNet} without using BN adaptation (Table \ref{table:gaussian} (a)).
However, adaptive BN for Adv$_2 [3]$ and Adv$_{\infty} [4/255]$ significantly improves the top-1 accuracy to $47.3\%$ and $44.9\%$ respectively.
In contrast, the baseline model only achieves $7.7\%$ accuracy.
We also observe similar results for {CIFAR-10} in Table \ref{table:gaussian} (b).

\smallskip
\begin{figure*}[h]
\centering
\begin{subfigure}[t]{0.11\linewidth}
\centering
\includegraphics[width=0.99\linewidth, height=45pt]{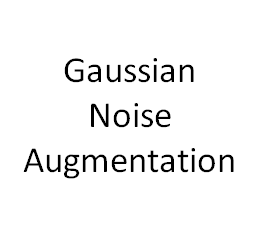} \\ \vspace{0.7em}
\includegraphics[width=0.99\linewidth, height=45pt]{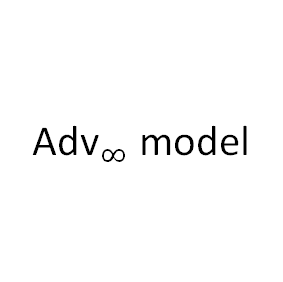} \\ \vspace{0.1em} 
\includegraphics[width=0.99\linewidth, height=45pt]{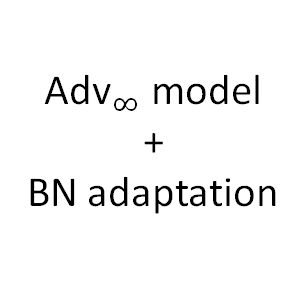} \\ \vspace{0.7em}
\includegraphics[width=0.99\linewidth, height=45pt]{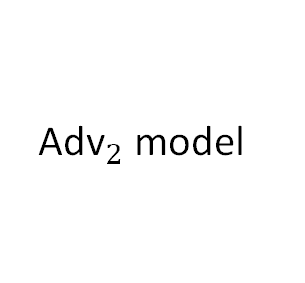} \\ \vspace{0.1em} \includegraphics[width=0.9\linewidth, height=35pt]{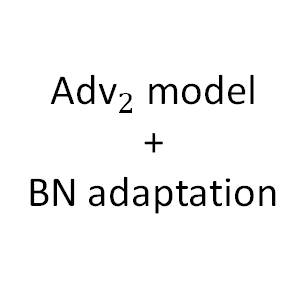} \\ 			
\end{subfigure}
\begin{subfigure}[t]{0.2\linewidth}
\centering
\includegraphics[width=0.99\linewidth, height=45pt]{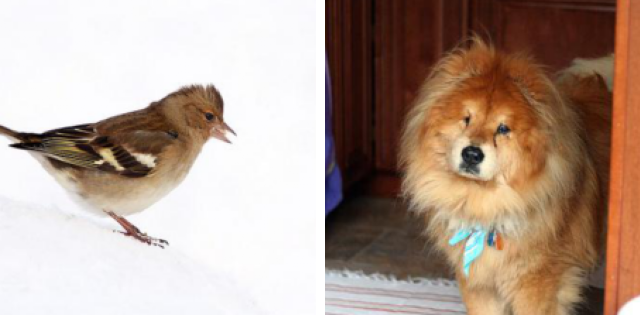} \\ \vspace{0.7em}
\includegraphics[width=0.99\linewidth, height=45pt]{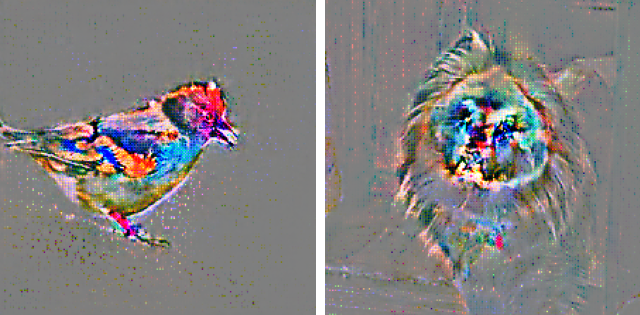} \\ \vspace{0.1em} 
\includegraphics[width=0.99\linewidth, height=45pt]{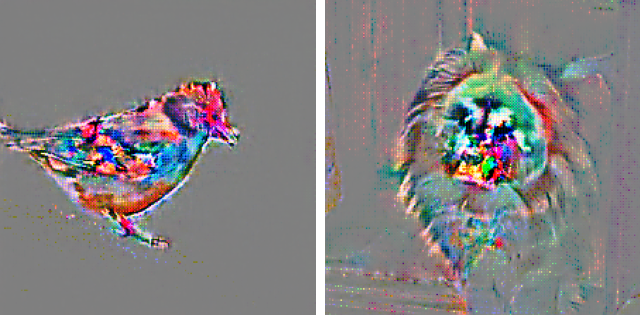} \\ \vspace{0.7em}
\includegraphics[width=0.99\linewidth, height=45pt]{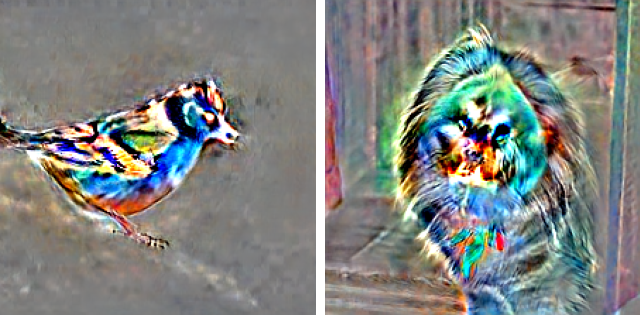} \\ \vspace{0.1em} 
\includegraphics[width=0.99\linewidth, height=45pt]{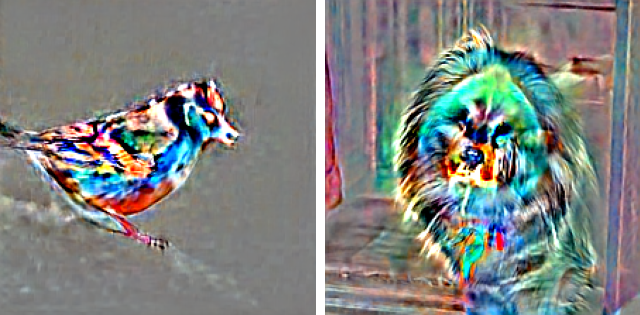} \\ 			
\caption{$\sigma=0$ (Clean)}
\label{fig_lossGrad_sigma0}
\end{subfigure}
\begin{subfigure}[t]{0.2\linewidth}
\centering
\includegraphics[width=0.99\linewidth, height=45pt]{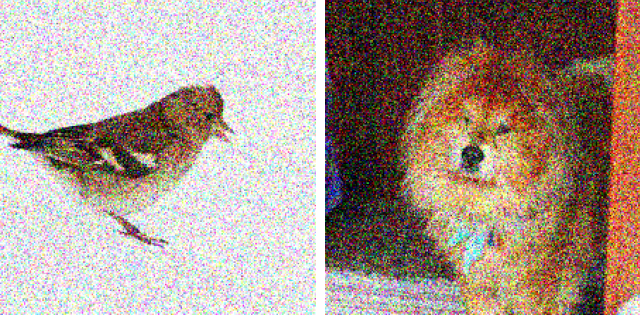} \\ \vspace{0.7em}
\includegraphics[width=0.99\linewidth, height=45pt]{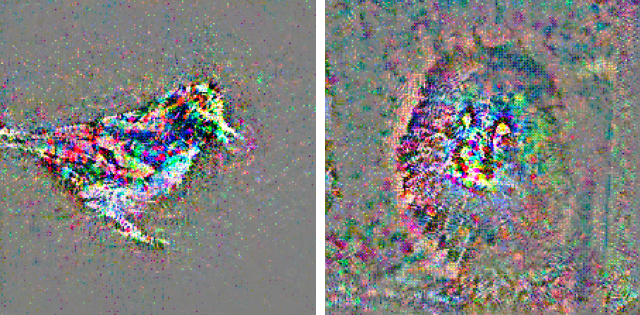} \\ \vspace{0.1em} 
\includegraphics[width=0.99\linewidth, height=45pt]{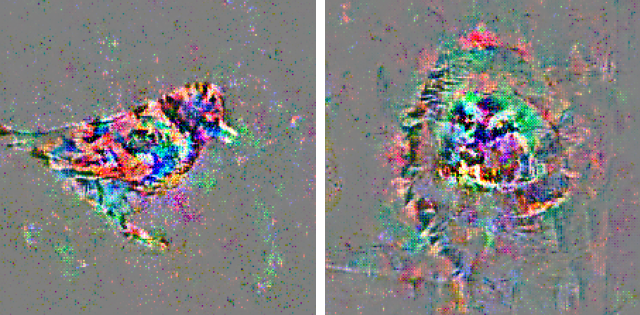} \\ \vspace{0.7em}
\includegraphics[width=0.99\linewidth, height=45pt]{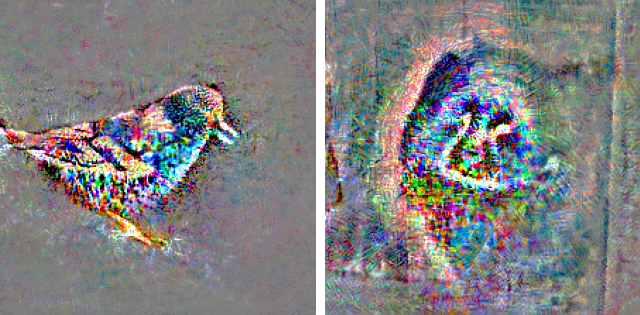} \\ \vspace{0.1em} 
\includegraphics[width=0.99\linewidth, height=45pt]{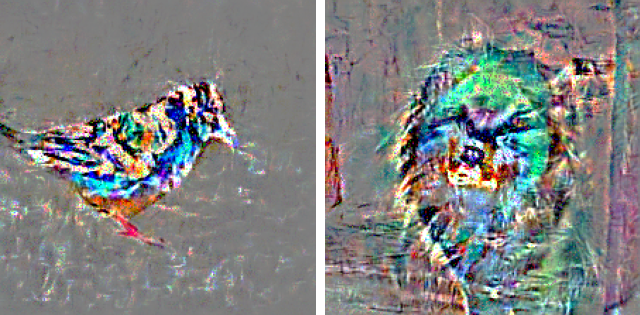} \\ 			
\caption{$\sigma=0.25$}
\label{fig_lossGrad_sigma25}
\end{subfigure}
\begin{subfigure}[t]{0.2\linewidth}
\centering
\includegraphics[width=0.99\linewidth, height=45pt]{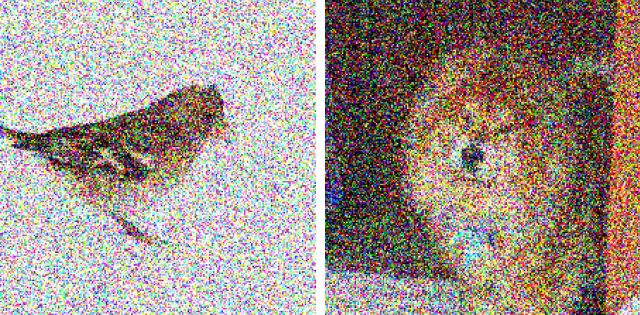} \\ \vspace{0.7em}
\includegraphics[width=0.99\linewidth, height=45pt]{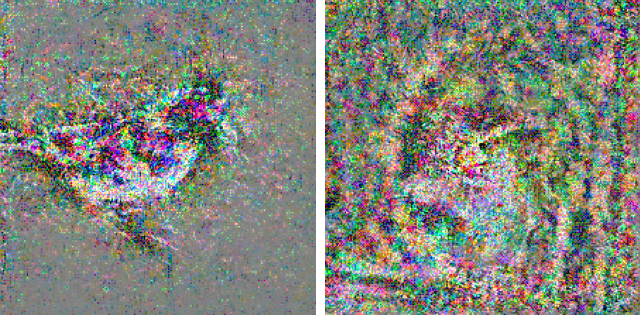} \\ \vspace{0.1em} 
\includegraphics[width=0.99\linewidth, height=45pt]{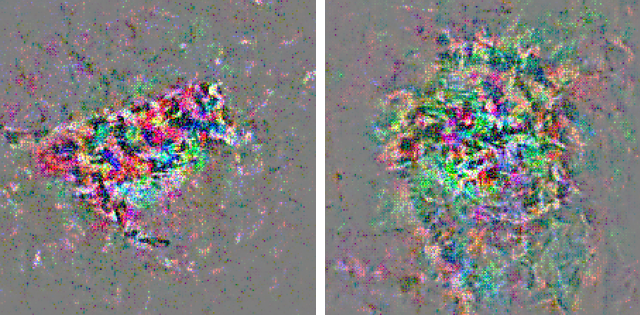} \\ \vspace{0.7em}
\includegraphics[width=0.99\linewidth, height=45pt]{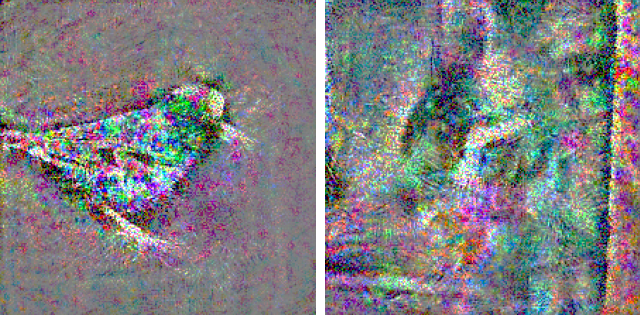} \\ \vspace{0.1em} 
\includegraphics[width=0.99\linewidth, height=45pt]{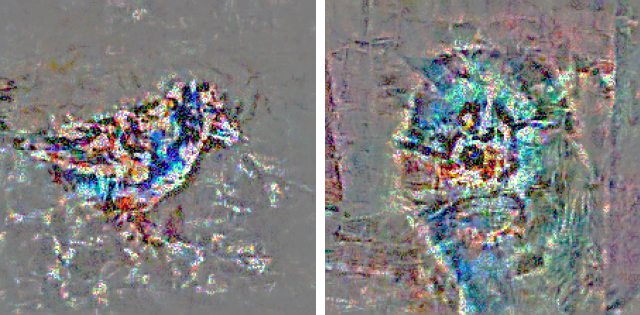} \\ 
\caption{$\sigma=0.5$}
\label{fig_lossGrad_sigma50}
\end{subfigure}
\begin{subfigure}[t]{0.2\linewidth}
\centering
\includegraphics[width=0.99\linewidth, height=45pt]{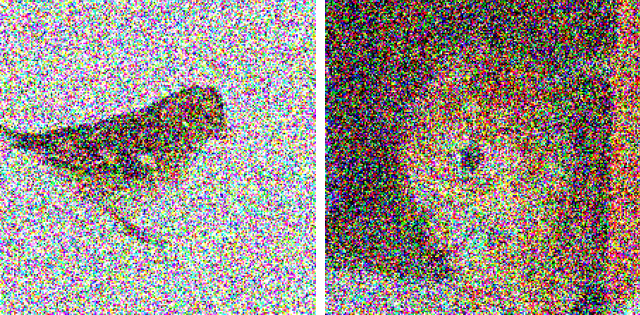} \\ \vspace{0.7em}
\includegraphics[width=0.99\linewidth, height=45pt]{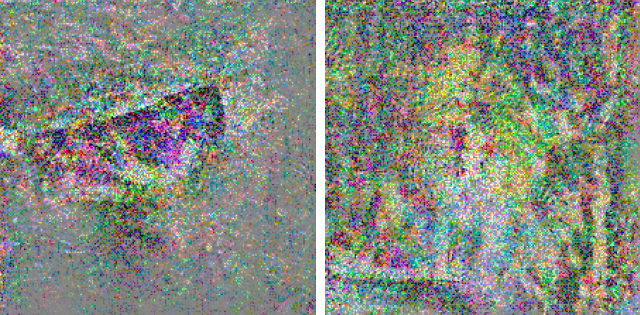} \\ \vspace{0.1em} 
\includegraphics[width=0.99\linewidth, height=45pt]{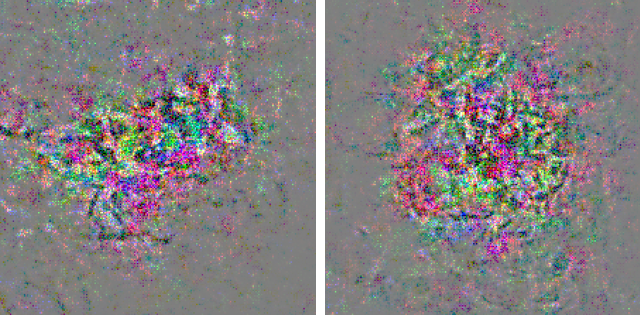} \\ \vspace{0.7em}
\includegraphics[width=0.99\linewidth, height=45pt]{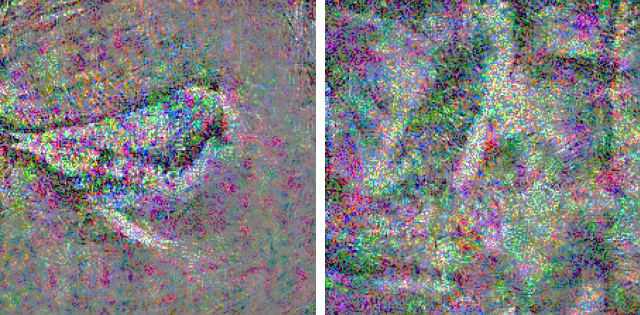} \\ \vspace{0.1em} 
\includegraphics[width=0.99\linewidth, height=45pt]{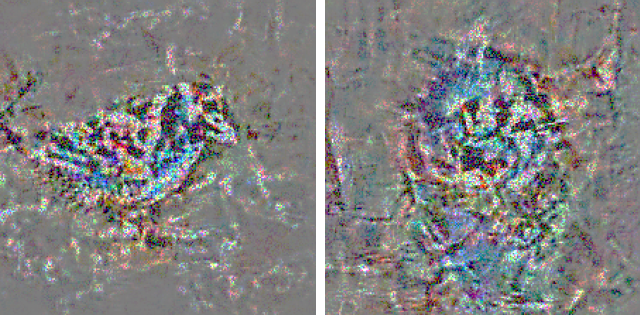} \\ 

\caption{$\sigma=0.75$}
\label{fig_lossGrad_sigma75}
\end{subfigure}
\vspace{-0.5em}
\caption{ Visualizing loss-gradients produced by AT models as we apply different levels of Gaussian noises. 
}
\label{fig:loss_gradients}
\end{figure*}

\smallskip
\textbf{Adaptive BN for AT models correctly extracts robust features under Gaussian noises. }
In Figure \ref{fig:loss_gradients}, we further investigate the performance of AT models by visualizing the \textit{loss gradients} of individual pixels of an image as we increase the noise level , $\sigma$.
\textit{Loss-gradients} reflect the most relevant input pixels for classification predictions.
Here, we scale, translate and clip the loss-gradient values without using any sophisticated techniques (as in \cite{robustnessDrops_iclr_2018}).
At $\sigma=0$ (i.e., for clean images), the loss-gradients from AT models align properly with perceptually relevant features (as observed previously \citep{robustnessDrops_iclr_2018,saliency_icml_2019}).
However, as we choose higher noise using $\sigma$=$0.5$ and $\sigma$=$0.75$, the overall loss gradients become noisier.
Specifically, we can see that AT models without BN adaptation produce larger gradient values (i.e., greater importance) even for background pixels.
In contrast, AT models with BN adaptation using Gaussian noises allows to correctly extract perceptually relevant features from the object of interest, suppressing the gradients for background (refer to Figure \ref{fig:loss_gradients}(c) and Figure \ref{fig:loss_gradients}(d)).
In other words, it allows us to extract the required semantic information for correct classifications.
Also, it is interesting to note that Adv$_2$ produces significantly more human-aligned loss gradients compared to Adv$_{\infty}$.
This behavior is also reflected in their classification performance in Table \ref{table:gaussian} and certification robustness in Table \ref{table_cifar10_certify}. 
We can see that Adv$_{2}$ overall produces better performance compared to Adv$_{\infty}$.
These results indicate that we can achieve non-trivial certification results by appropriately adapting the AT models, as demonstrated in the following sections.

\subsection{Certified Robustness for AT models}
We now present the $\ell_2$ certification results using the randomized smoothing framework as the backbone, as proposed in our Algorithm \ref{algo_certify}.
We certify the test images with $99.9\%$ probability.
We estimate the class-label probabilities of $g$ (Equation \ref{eq_certify_l2}) using $100,000$ noisy samples, as in \citep{certiSmoothing_icml_2019,advSmooth_nips_2019}.
We use the full test-set of $10,000$ images for {CIFAR-10} and a sub-sample of $500$ test images for {ImageNet}.

\begin{figure}[h]
\centering
\begin{subfigure}[t]{0.49\linewidth}
\centering
\includegraphics[width=0.8\linewidth, height=110pt]{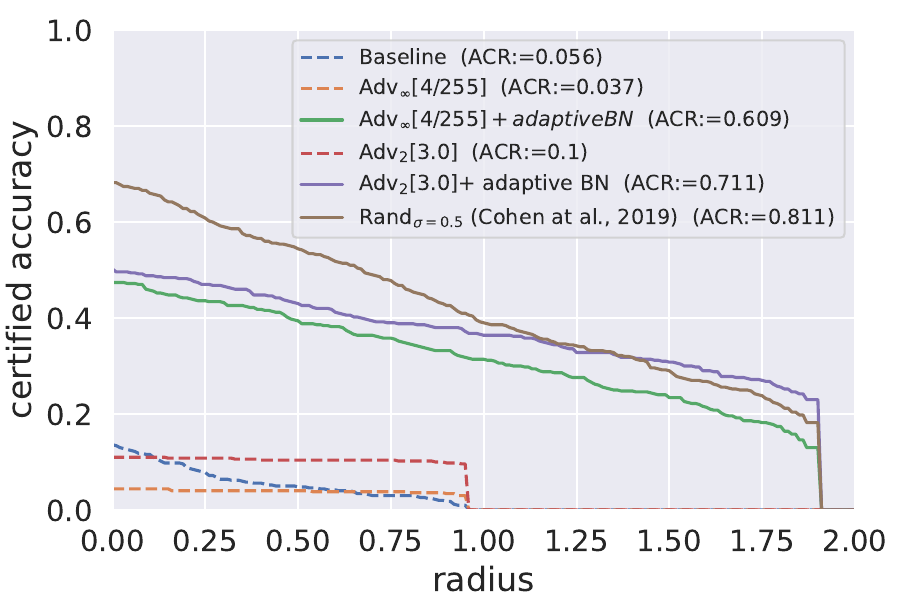}
\end{subfigure}
\begin{subfigure}[t]{0.49\linewidth}
\centering
\includegraphics[width=0.8\linewidth, height=110pt]{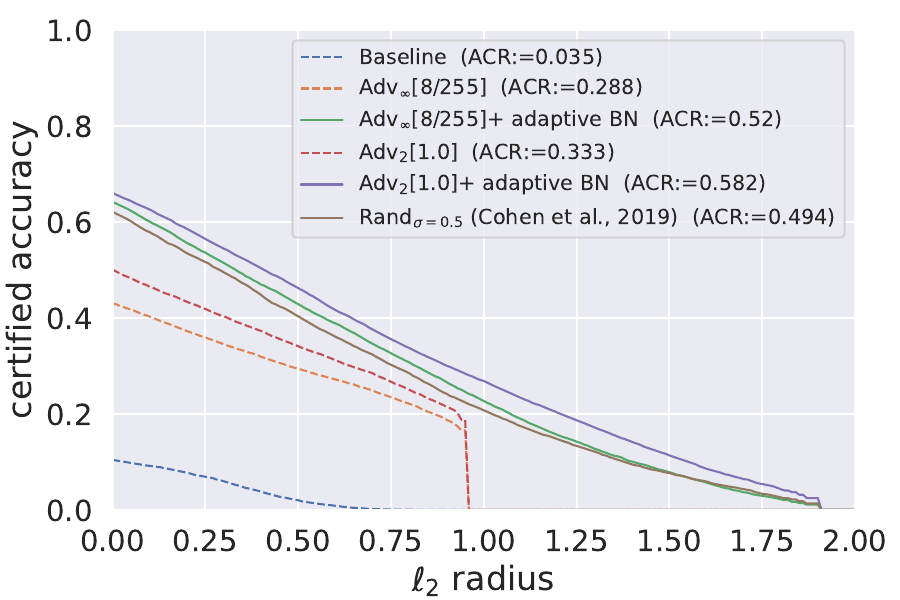}
\end{subfigure}
\vspace{-0.5em}
\caption{ Certification through adaptation produces non-trivial certified robustness at various $\ell_2$ radii for both \textbf{(Left)} {ImageNet} and \textbf{(Right)} {CIFAR-10} datasets.}
\label{fig:certify}
\end{figure}

\textbf{Non-trivial certification for AT models. }
In Figure \ref{fig:certify}, we first demonstrate that AT models can achieve non-trivial $\ell_2$ certified robustness using our proposed certification through adaptation technique for both {ImageNet} and {CIFAR-10} datasets.
Here, we use Adv$_{\infty}[4/255]$ and Adv$_2[3]$ for ImageNet and Adv$_{\infty}[8/255]$ and Adv$_2[1]$ for CIFAR-10. 
We apply fixed noise levels of $\sigma=0.5$ to certify all test examples using proposed Algorithm \ref{algo_certify}.
Here, we compare with the certification results of the Baseline, Adv$_{\infty}$ and Adv$_2$ models at $\sigma=0.25$ in the standard settings (i.e., without adapting these models).
We can see a significant boost of $\ell_2$ certification results for both Adv$_{\infty}$ and Adv$_2$ models using our proposed framework.
Further, Adv$_2$ models consistently achieve better certification performance compared to Adv$_{\infty}$.
We also compare with Rand$_{\sigma=0.5}$ models at fixed $\sigma=0.5$, as in \cite{certiSmoothing_icml_2019}.
For {CIFAR-10}, both Adv$_{\infty}[8/255]$ and Adv$_2[1]$ outperform the  Rand$_{\sigma=0.5}$ models \citep{certiSmoothing_icml_2019}.
Furthermore, for {ImageNet}, Adv$_2[3]$ achieves better certified accuracy compared to Rand$_{\sigma=0.5}$ beyond $\ell_2$-radii of $1.5$. 
Please refer to Table \ref{table_app_imagenet_certify} and \ref{table_app_cifar10_certify} (Appendix) for detailed comparisons of different models, trained using different specifications.

\begin{figure*}[h]
\centering
\begin{subfigure}[t]{0.32\linewidth}
\centering
\includegraphics[width=1\linewidth, height=110pt]{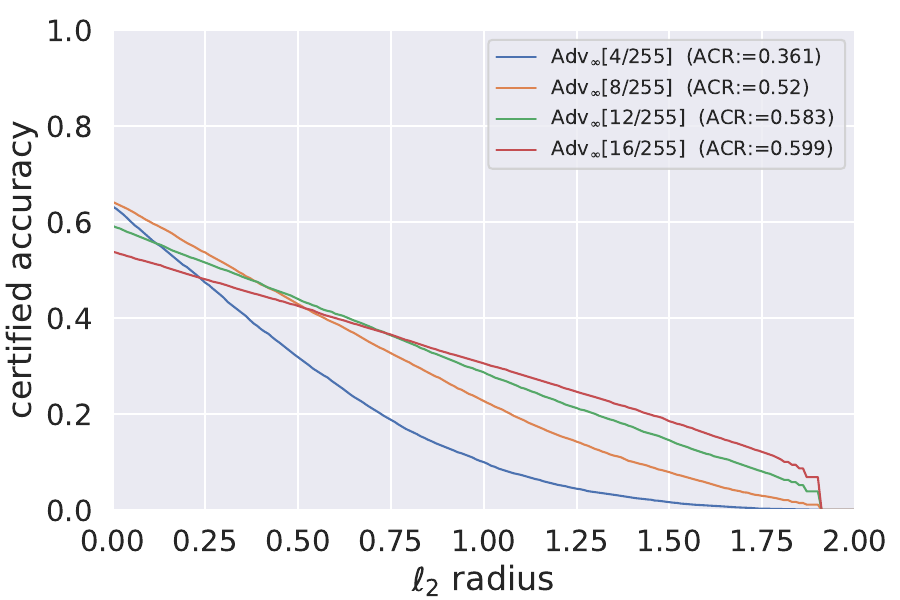}
\caption{}
\end{subfigure}
\begin{subfigure}[t]{0.32\linewidth}
\centering
\includegraphics[width=1\linewidth, height=110pt]{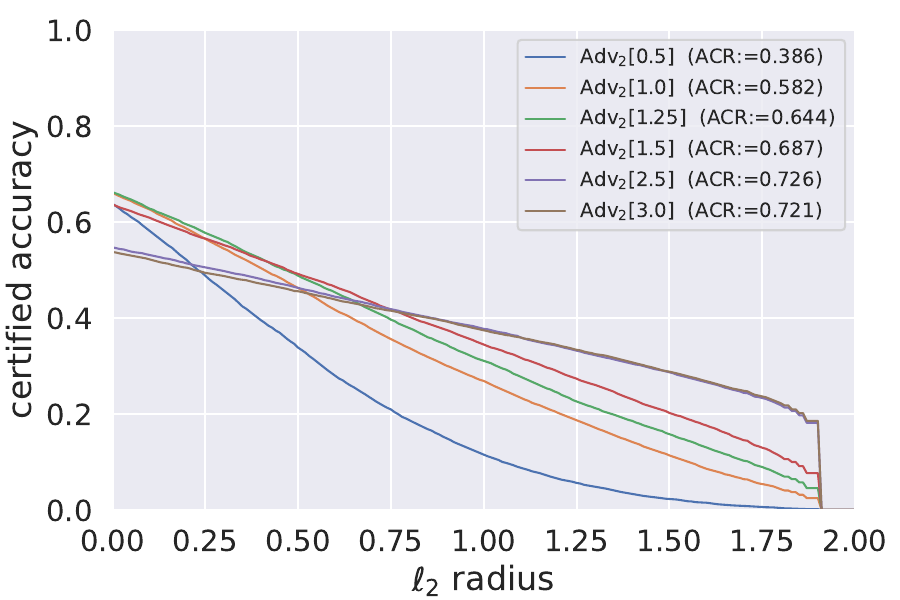}
\caption{}
\end{subfigure}
\begin{subfigure}[t]{0.32\linewidth}
\centering
\includegraphics[width=1\linewidth, height=110pt]{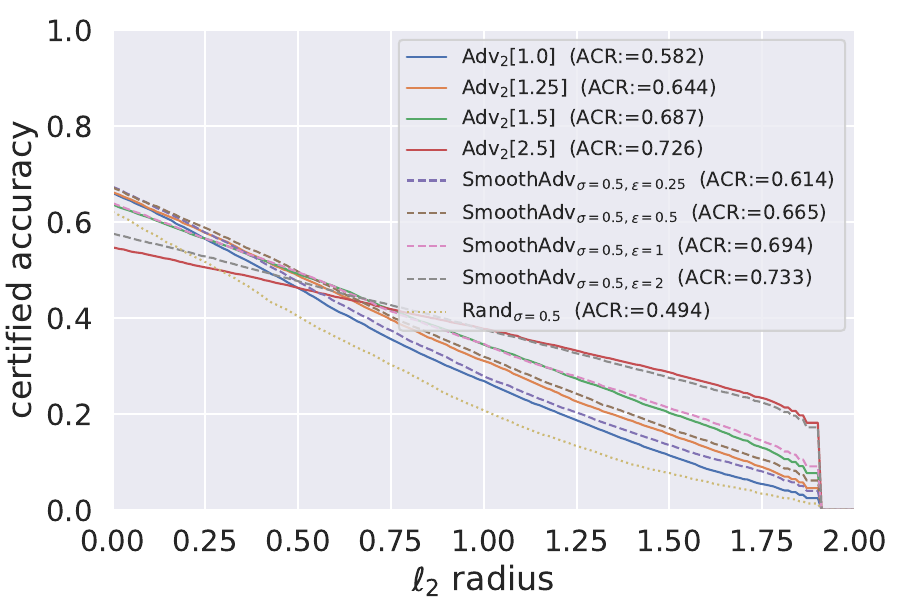}
\caption{}
\end{subfigure}
\vspace{-0.5em}
\caption{ {CIFAR-10}:  AT models, trained using larger threat boundaries, produces better certification results for higher $\ell_2$ radii.
Here, we apply our proposed certification through adaptation with fixed noise-level $\sigma=0.5$.
(c) Comparison of Adv$_2$ models with SmoothAdv \citep{advSmooth_nips_2019}, trained and certified using $\sigma=0.5$.
}
\label{fig:certify_varyBoundary}
\end{figure*}
\medskip
\textbf{Larger Threat Boundary for Better Certification. }
Learning AT models at a higher threat boundary produces better certified robustness at higher $\ell_2$ radii.
Figure \ref{fig:certify_varyBoundary}(a) and \ref{fig:certify_varyBoundary}(b) demonstrate this phenomena for {CIFAR-10} on both Adv$_{\infty}$ and Adv$_2$ models respectively .

Figure \ref{fig:certify_varyBoundary}(c) also compares the certified accuracy of Adv$_2$ models with the existing state-of-the-art \textit{SmoothAdv} models \citep{advSmooth_nips_2019}.
SmoothAdv utilizes adversarial training using an adaptive attack with an $\ell_2$ threat boundary of $\epsilon$ and Gaussian noises, $\mathcal{N}(0, \sigma^2I)$ (See details in Appendix \ref{sec_implementation}).
We set the noise to $\sigma=0.5$ and vary $\epsilon$ for their training to compare with different SmoothAdv models in Figure \ref{fig:certify_varyBoundary}(c).
As we can see that by adapting  Adv$_2$ models at $\sigma=0.5$ using our proposed Algorithm \ref{algo_certify}, we can already achieve similar performance as SmoothAdv.
Next, we demonstrate that our proposed \textit{Auto-Noise} technique further improves the performance of both AT models and existing randomized smoothing based models.

\begin{figure*}[h]
\centering
\begin{subfigure}[t]{0.31\linewidth}
\centering
\includegraphics[width=1\linewidth, height=110pt]{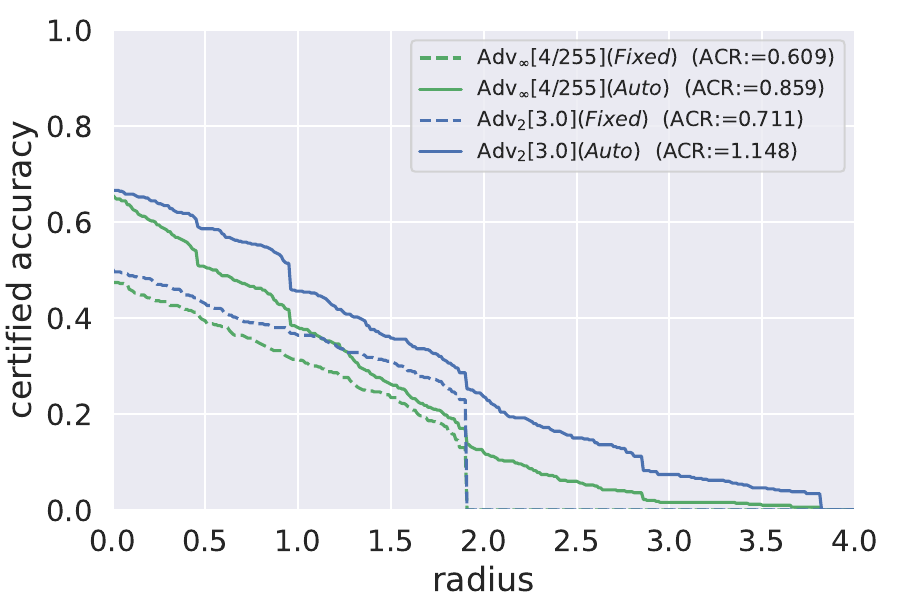}
\caption{ {ImageNet}}
\end{subfigure}
\begin{subfigure}[t]{0.31\linewidth}
\centering
\includegraphics[width=1\linewidth, height=110pt]{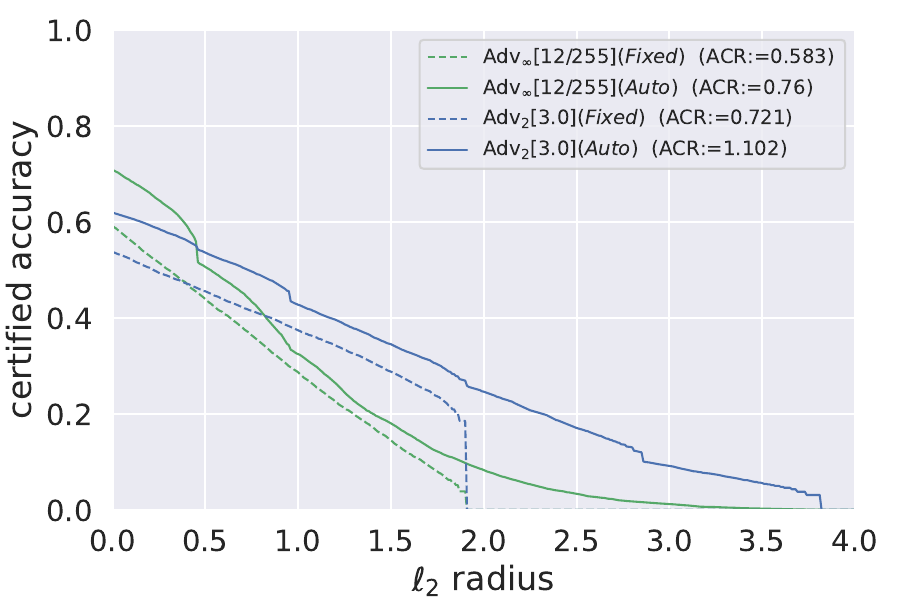}
\caption{ {CIFAR-10}}
\end{subfigure}
\begin{subfigure}[t]{0.35\linewidth}
\centering
\includegraphics[width=1\linewidth, height=110pt]{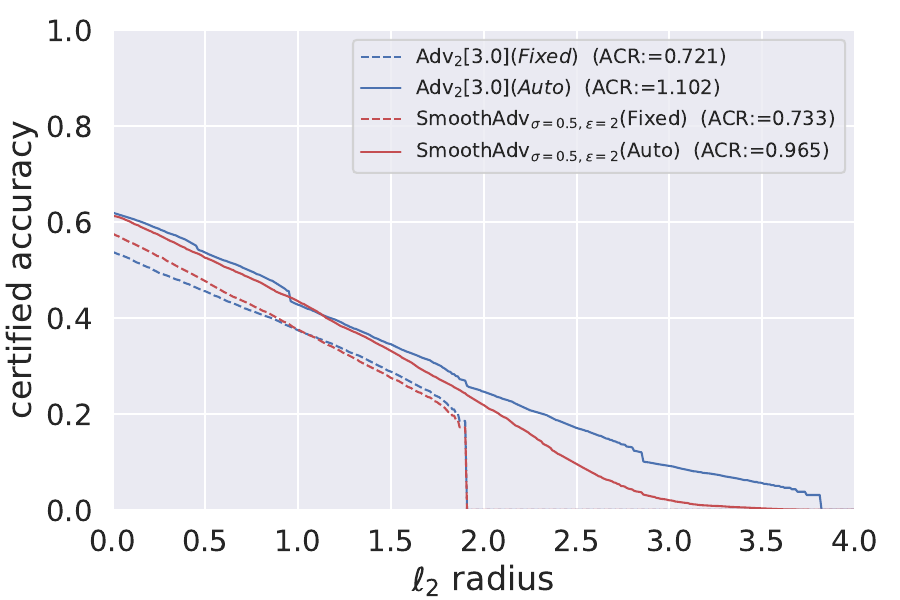}
\caption{ SmoothAdv  vs Adv$_2$ (CIFAR-10)}
\end{subfigure}
\caption{ Auto-Noise (denoted as "auto") vs. fixed noise at $\sigma=0.5$ (denoted as "fixed") for $\ell_2$ certification on (a) {ImageNet} and (b) {CIFAR-10} datasets for AT models.
(c) SmoothAdv  vs Adv$_2$ models for {CIFAR-10} using Auto-Noise technique.
Here, we only present the results with the best ACR scores.
}
\label{fig:certify_varyNoise}
\end{figure*}

\subsection{Auto-Noise: Flexibility of choosing appropriate $\sigma$ for certification.}
\label{sec_autonoise_experiment}
In Table \ref{table:gaussian}, we can see that the classification models remain robust only for a few test examples under higher Gaussian noise. 
It suggests that the optimal noise levels for certifying different test examples may vary significantly, indicating the importance of our proposed Auto-Noise technique.

\textbf{Auto-Noise for AT models. }
Figure \ref{fig:certify_varyNoise}(a) and \ref{fig:certify_varyNoise}(b) present the certification performance of AT models as we apply both certification through adaptation (Algorithm~\ref{algo_certify}) and Auto-Noise technique for for ImageNet and CIFAR-10 datasets respectively.
We compare their performance by certifying using a fixed noise level, $\sigma=0.5$.
We can see that the Auto-Noise technique can significantly improve the performance to achieve $1.148$ and $1.102$ ACR scores for the best AT models on ImageNet and CIFAR-10 datasets.
Figure \ref{fig:certify_varyNoise}(c) demonstrates that the Auto-Noise technique also improves the performance of the best SmoothAdv model to an ACR score of $0.965$ for CIFAR-10. 
However, our \emph{Certification through Adaptation} together with \emph{Auto-Noise} technique for Adv$_2[3]$ model outperforms the best SmoothAdv model for CIFAR-10.

\begin{figure*}[h]
\centering
\begin{subfigure}[t]{0.32\linewidth}
\centering
\includegraphics[width=1\linewidth, height=110pt]{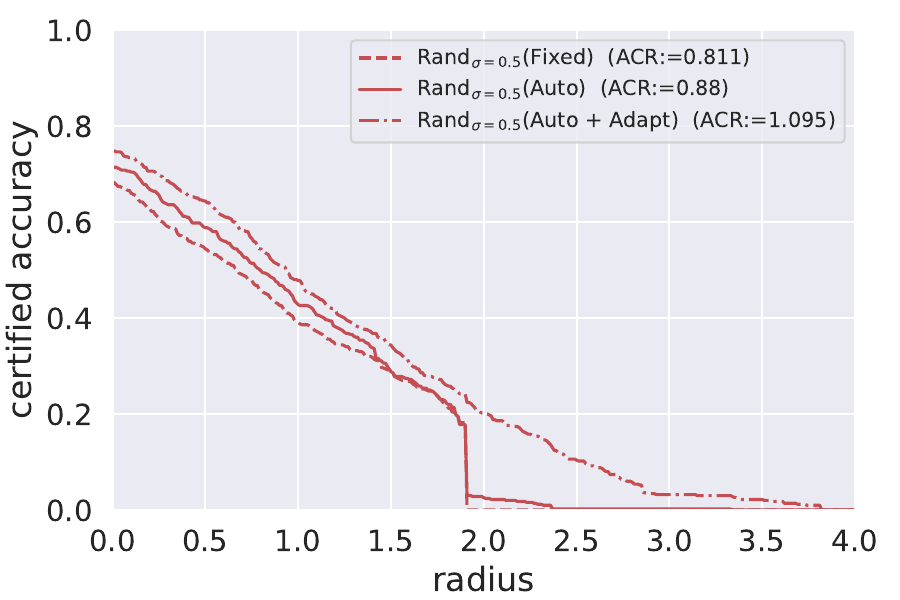}
\caption{ Rand$_{\sigma=0.5}$ ({ImageNet})}
\end{subfigure}
\begin{subfigure}[t]{0.32\linewidth}
\centering
\includegraphics[width=1\linewidth, height=110pt]{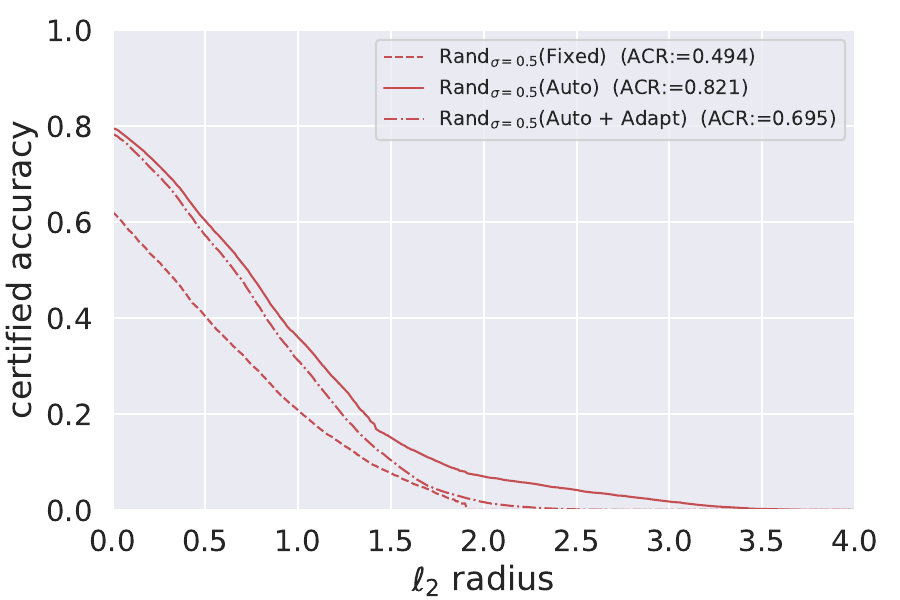}
\caption{ Rand$_{\sigma=0.5}$ ({CIFAR-10})}
\end{subfigure}
\begin{subfigure}[t]{0.32\linewidth}
\centering
\includegraphics[width=1\linewidth, height=110pt]{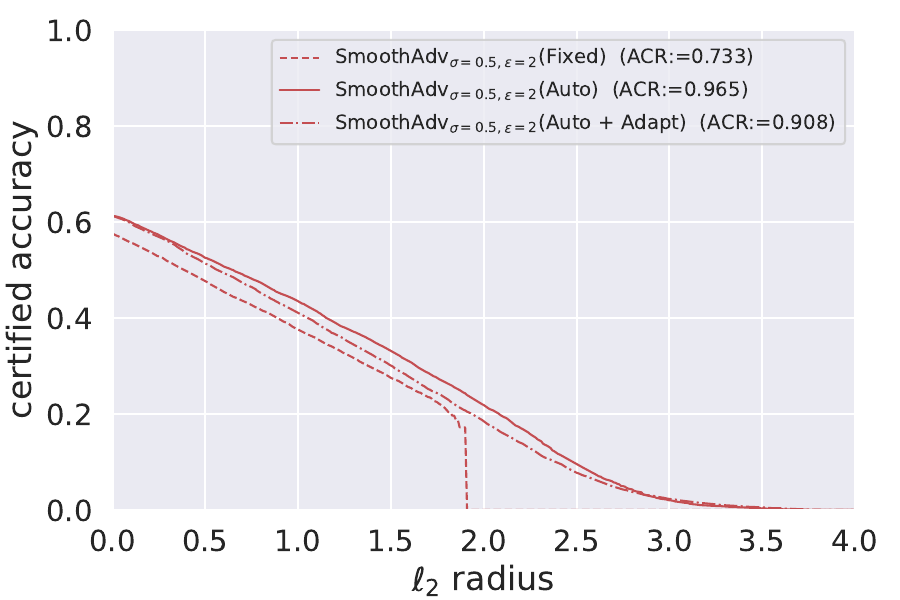}
\caption{ SmoothAdv ({CIFAR-10})}
\end{subfigure}
\caption{Effect of adaptation and Auto-Noise technique for existing randomized smoothing based models.}
\label{fig:certify_autoNoise_rs}
\end{figure*}

\textbf{Certification through Adaptation along with Auto-Noise for existing models.}
In Figure \ref{fig:certify_autoNoise_rs}, we compare the performance of existing randomized smoothing based models as we adapt their base models using Algorithm \ref{algo_certify} followed by Auto-Noise technique.
We can see in Figure \ref{fig:certify_autoNoise_rs}(a) that incorporating adaptation using Algorithm \ref{algo_certify} improves the Rand$_{\sigma=0.5}$ model to achieve ACR score of  $1.095$ on ImageNet dataset.
In contrast, the certification performance for both Rand$_{\sigma=0.5}$ and SmoothAdv models degrades for CIFAR-10 datasets (Figure \ref{fig:certify_autoNoise_rs}(b) and Figure \ref{fig:certify_autoNoise_rs}(c)).
Hence, we do not include these models in Figure \ref{fig:certify_varyNoise} (c) to compare with AT models.

\begin{figure}[h]
\centering
\begin{subfigure}[t]{0.49\linewidth}
\centering
\includegraphics[width=0.8\linewidth, height=110pt]{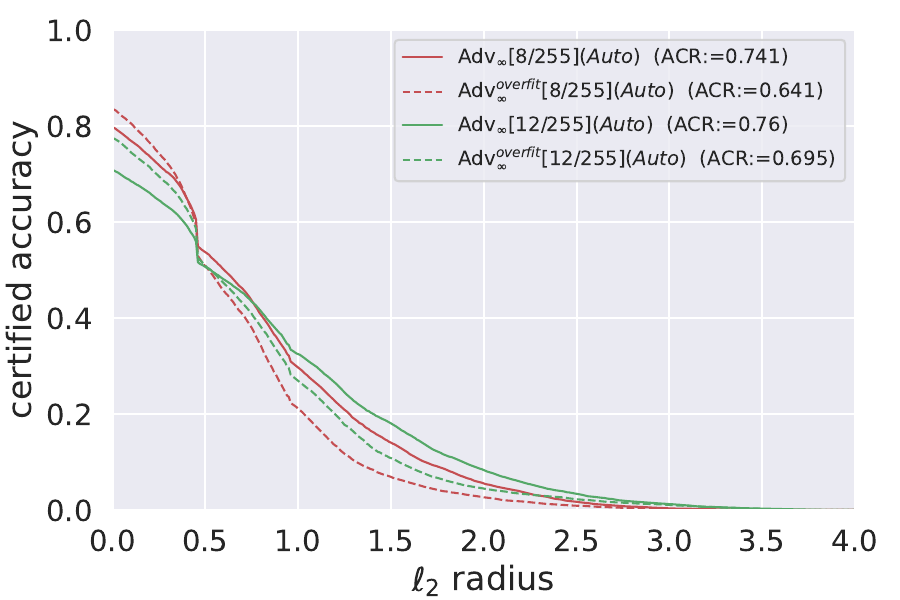}
\caption{Adv$_{\infty}$ models}
\end{subfigure}
\begin{subfigure}[t]{0.49\linewidth}
\centering
\includegraphics[width=0.8\linewidth, height=110pt]{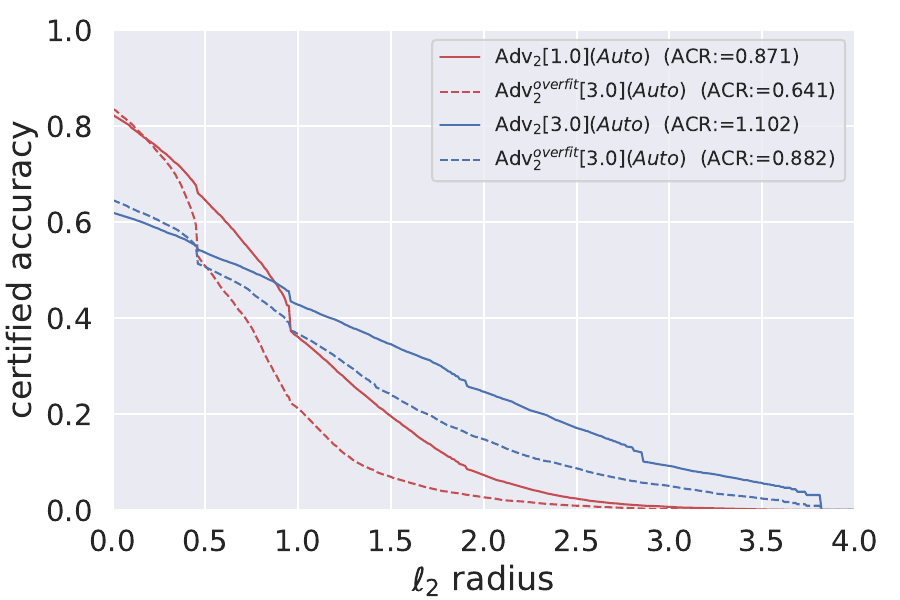}
\caption{Adv$_{2}$ models}
\end{subfigure}
\caption{{CIFAR-10}:  Certification performance degrades for over-fitted AT models when trained without applying early-stopping criteria \citep{advOverfitting_icml_2020}. The over-fitted models are denoted as Adv$^{overfit}$.
}
\label{fig:certify_overfit}\vspace{-0.5em}
\end{figure}

\subsection{Over-fitted AT models degrades certification}
\cite{advOverfitting_icml_2020} demonstrate that AT models \textit{overfit} when trained without \textit{early stopping} criteria.
It degrades their empirical robustness against adversarial attacks.
Figure \ref{fig:certify_overfit} compares with the certification results of such \textit{overfitted} AT models, denoted as Adv$^{overfit}$.
We observe that Adv$^{overfit}$ models also degrade the certified robustness compared to their corresponding AT models, trained with early stopping criteria.
In particular, the difference in their certification performance is more prominent at higher $\ell_2$ radii.
Hence, these results as well as our results in Figure \ref{fig:certify_varyNoise} indicate that empirical and certified robustness are closely related. 
In particular, \textit{improving the empirical robustness for a model also allows to provide better certified robustness.}

\section{Conclusion}
We propose a novel \textit{certification through adaptation} algorithm that transforms the AT models into a randomized smoothing classifier to provide certified robustness for $\ell_2$ norm.
We also propose \emph{Auto-Noise} to efficiently approximate the appropriate noise levels to certify different test examples.
Empirically we improve the performance of both AT models and existing randomized smoothing-based models on CIFAR-10 and ImageNet datasets using the Auto-Noise technique.
Further, our \emph{Certification through Adaptation} together with \emph{Auto-Noise} technique significantly improves the ACR scores using AT models.
Notably, our framework does not affect the empirical robustness or benign accuracy of an AT model to provide these non-trivial $\ell_2$ certification results.
Hence, our paper is a step towards bridging the gap between current research of empirical and certified robustness against adversarial examples.

\subsubsection*{Broader Impact Statement}
Improving empirical robustness against adversarial examples along with certification guarantees is an important problem to enhance the reliability of a DNN model for sensitive real-world applications.
However, the current state-of-the-art defense methods against adversarial examples typically focus on improving either empirical or certified robustness.
In this paper, we aim to bridge this gap by significantly improving the certification performance of AT models without affecting the benign accuracy or reducing their state-of-the-art empirical robustness.



\bibliography{example_paper}

\begin{thebibliography}{79}
\providecommand{\natexlab}[1]{#1}
\providecommand{\url}[1]{\texttt{#1}}
\expandafter\ifx\csname urlstyle\endcsname\relax
  \providecommand{\doi}[1]{doi: #1}\else
  \providecommand{\doi}{doi: \begingroup \urlstyle{rm}\Url}\fi

\bibitem[Athalye et~al.(2018)Athalye, Carlini, and
  Wagner]{obfuscated_icml_2018}
Anish Athalye, Nicholas Carlini, and David Wagner.
\newblock Obfuscated gradients give a false sense of security: Circumventing
  defenses to adversarial examples.
\newblock In \emph{ICML}, 2018.

\bibitem[Benz et~al.(2021{\natexlab{a}})Benz, Zhang, Karjauv, and
  Kweon]{adaptBN_cp_wacv_2021}
Philipp Benz, Chaoning Zhang, Adil Karjauv, and In~So Kweon.
\newblock Revisiting batch normalization for improving corruption robustness.
\newblock In \emph{WACV}, 2021{\natexlab{a}}.

\bibitem[Benz et~al.(2021{\natexlab{b}})Benz, Zhang, and
  Kweon]{bnreduce_advrobust_iccv21}
Philipp Benz, Chaoning Zhang, and In~So Kweon.
\newblock Batch normalization increases adversarial vulnerability and decreases
  adversarial transferability: A non-robust feature perspective.
\newblock In \emph{ICCV}, 2021{\natexlab{b}}.

\bibitem[Bunel et~al.(2018)Bunel, Turkaslan, Torr, Kohli, and
  Kumar]{bunel_nips_2018}
Rudy Bunel, Ilker Turkaslan, Philip~HS Torr, Pushmeet Kohli, and M~Pawan Kumar.
\newblock A unified view of piecewise linear neural network verification.
\newblock \emph{NeurIPS}, 2018.

\bibitem[Cao \& Gong(2017)Cao and Gong]{smooth1_heu_2017}
Xiaoyu Cao and Neil~Zhenqiang Gong.
\newblock Mitigating evasion attacks to deep neural networks via region-based
  classification.
\newblock In \emph{ACSAC}, 2017.

\bibitem[Cariucci et~al.(2017)Cariucci, Porzi, Caputo, Ricci, and
  Bulo]{cariucci2017autodial}
Fabio~Maria Cariucci, Lorenzo Porzi, Barbara Caputo, Elisa Ricci, and
  Samuel~Rota Bulo.
\newblock Autodial: Automatic domain alignment layers.
\newblock In \emph{ICCV}, 2017.

\bibitem[Carlini \& Wagner(2017)Carlini and Wagner]{cwAttack_sp_2017}
N.~Carlini and D.~Wagner.
\newblock Towards evaluating the robustness of neural networks.
\newblock In \emph{IEEE S\&P}, 2017.

\bibitem[Carlini et~al.(2022)Carlini, Tramer, Kolter,
  et~al.]{carlini2022certified}
Nicholas Carlini, Florian Tramer, J~Zico Kolter, et~al.
\newblock (certified!!) adversarial robustness for free!
\newblock \emph{arXiv preprint arXiv:2206.10550}, 2022.

\bibitem[Carmon et~al.(2019)Carmon, Raghunathan, Schmidt, Liang, and
  Duchi]{advUnlabelled1_arxiv_2019}
Yair Carmon, Aditi Raghunathan, Ludwig Schmidt, Percy Liang, and John~C Duchi.
\newblock Unlabeled data improves adversarial robustness.
\newblock \emph{arXiv}, 2019.

\bibitem[Cohen et~al.(2019)Cohen, Rosenfeld, and
  Kolter]{certiSmoothing_icml_2019}
Jeremy~M Cohen, Elan Rosenfeld, and J~Zico Kolter.
\newblock Certified adversarial robustness via randomized smoothing.
\newblock \emph{ICML}, 2019.

\bibitem[Deng et~al.(2009)Deng, Dong, Socher, Li, Li, and Fei-Fei]{db_imagenet}
Jia Deng, Wei Dong, Richard Socher, Li-Jia Li, Kai Li, and Li~Fei-Fei.
\newblock Imagenet: A large-scale hierarchical image database.
\newblock In \emph{CVPR}, 2009.

\bibitem[Dvijotham et~al.(2018)Dvijotham, Stanforth, Gowal, Mann, and
  Kohli]{certify10_uai_2018}
Krishnamurthy Dvijotham, Robert Stanforth, Sven Gowal, Timothy~A Mann, and
  Pushmeet Kohli.
\newblock A dual approach to scalable verification of deep networks.
\newblock In \emph{UAI}, 2018.

\bibitem[Dvijotham et~al.(2020)Dvijotham, Hayes, Balle, Kolter, Qin, Gyorgy,
  Xiao, Gowal, and Kohli]{certiLP_iclr_2020}
Krishnamurthy~(Dj) Dvijotham, Jamie Hayes, Borja Balle, Zico Kolter, Chongli
  Qin, Andras Gyorgy, Kai Xiao, Sven Gowal, and Pushmeet Kohli.
\newblock A framework for robustness certification of smoothed classifiers
  using f-divergences.
\newblock In \emph{ICLR}, 2020.

\bibitem[Engstrom et~al.(2019)Engstrom, Ilyas, Salman, Santurkar, and
  Tsipras]{advImageNetModels_2019}
Logan Engstrom, Andrew Ilyas, Hadi Salman, Shibani Santurkar, and Dimitris
  Tsipras.
\newblock Robustness (python library), 2019.
\newblock URL \url{https://github.com/MadryLab/robustness}.

\bibitem[Etmann et~al.(2019)Etmann, Lunz, Maass, and
  Sch{\"o}nlieb]{saliency_icml_2019}
Christian Etmann, Sebastian Lunz, Peter Maass, and Carola-Bibiane
  Sch{\"o}nlieb.
\newblock On the connection between adversarial robustness and saliency map
  interpretability.
\newblock In \emph{ICML}, 2019.

\bibitem[French et~al.(2017)French, Mackiewicz, and Fisher]{adaptation2_2017}
Geoffrey French, Michal Mackiewicz, and Mark Fisher.
\newblock Self-ensembling for visual domain adaptation.
\newblock \emph{arXiv}, 2017.

\bibitem[Gehr et~al.(2018)Gehr, Mirman, Drachsler-Cohen, Tsankov, Chaudhuri,
  and Vechev]{certify17_sp_2018}
Timon Gehr, Matthew Mirman, Dana Drachsler-Cohen, Petar Tsankov, Swarat
  Chaudhuri, and Martin Vechev.
\newblock Ai2: Safety and robustness certification of neural networks with
  abstract interpretation.
\newblock In \emph{IEEE SP}, 2018.

\bibitem[Gilmer et~al.(2019)Gilmer, Ford, Carlini, and
  Cubuk]{cpGaussianAug_icml_2019}
Justin Gilmer, Nicolas Ford, Nicholas Carlini, and Ekin Cubuk.
\newblock Adversarial examples are a natural consequence of test error in
  noise.
\newblock In \emph{ICML}, 2019.

\bibitem[Goodfellow et~al.(2015)Goodfellow, Shlens, and
  Szegedy]{fgsm_iclr_2015}
Ian Goodfellow, Jonathon Shlens, and Christian Szegedy.
\newblock Explaining and harnessing adversarial examples.
\newblock In \emph{ICLR}, 2015.

\bibitem[Gowal et~al.(2018)Gowal, Dvijotham, Stanforth, Bunel, Qin, Uesato,
  Arandjelovic, Mann, and Kohli]{certify16_arxiv_2018}
Sven Gowal, Krishnamurthy Dvijotham, Robert Stanforth, Rudy Bunel, Chongli Qin,
  Jonathan Uesato, Relja Arandjelovic, Timothy Mann, and Pushmeet Kohli.
\newblock On the effectiveness of interval bound propagation for training
  verifiably robust models.
\newblock \emph{arXiv}, 2018.

\bibitem[Gowal et~al.(2020)Gowal, Qin, Uesato, Mann, and
  Kohli]{advTread2Gowal_2020}
Sven Gowal, Chongli Qin, Jonathan Uesato, Timothy Mann, and Pushmeet Kohli.
\newblock Unering the limits of adversarial training against norm-bounded
  adversarial examples.
\newblock \emph{arXiv}, 2020.

\bibitem[He et~al.(2016{\natexlab{a}})He, Zhang, Ren, and
  Sun]{resnet_cvpr_2016}
Kaiming He, Xiangyu Zhang, Shaoqing Ren, and Jian Sun.
\newblock Deep residual learning for image recognition.
\newblock In \emph{IEEE CVPR}, 2016{\natexlab{a}}.

\bibitem[He et~al.(2016{\natexlab{b}})He, Zhang, Ren, and
  Sun]{resnet_eccv_2016}
Kaiming He, Xiangyu Zhang, Shaoqing Ren, and Jian Sun.
\newblock Identity mappings in deep residual networks.
\newblock In \emph{ECCV}, 2016{\natexlab{b}}.

\bibitem[Hendrycks \& Dietterich(2019)Hendrycks and
  Dietterich]{commonPerturbation_iclr_2019}
Dan Hendrycks and Thomas Dietterich.
\newblock Benchmarking neural network robustness to common corruptions and
  perturbations.
\newblock In \emph{ICLR}, 2019.

\bibitem[Horv{\'a}th et~al.(2022)Horv{\'a}th, M{\"u}ller, Fischer, and
  Vechev]{boostingRS_iclr_2022}
Mikl{\'o}s~Z Horv{\'a}th, Mark~Niklas M{\"u}ller, Marc Fischer, and Martin
  Vechev.
\newblock Boosting randomized smoothing with variance reduced classifiers.
\newblock \emph{ICLR}, 2022.

\bibitem[Huang et~al.(2018)Huang, Yang, Lang, and Deng]{huang2018decorrelated}
Lei Huang, Dawei Yang, Bo~Lang, and Jia Deng.
\newblock Decorrelated batch normalization.
\newblock In \emph{CVPR}, 2018.

\bibitem[Ioffe \& Szegedy(2015)Ioffe and Szegedy]{bn_icml_2015}
Sergey Ioffe and Christian Szegedy.
\newblock Batch normalization: Accelerating deep network training by reducing
  internal covariate shift.
\newblock In \emph{ICML}, 2015.

\bibitem[Jalal et~al.(2019)Jalal, Ilyas, Daskalakis, and
  Dimakis]{overpoweredAttack_2017}
Ajil Jalal, Andrew Ilyas, Constantinos Daskalakis, and Alexandros~G Dimakis.
\newblock The robust manifold defense: Adversarial training using generative
  models.
\newblock \emph{arXiv}, 2019.

\bibitem[Jeong \& Shin(2020)Jeong and Shin]{consistency_nips_2020}
Jongheon Jeong and Jinwoo Shin.
\newblock Consistency regularization for certified robustness of smoothed
  classifiers.
\newblock \emph{NeurIPS}, 2020.

\bibitem[Jiang et~al.(2020)Jiang, Chen, Chen, and Wang]{dualBN_1}
Ziyu Jiang, Tianlong Chen, Ting Chen, and Zhangyang Wang.
\newblock Robust pre-training by adversarial contrastive learning.
\newblock \emph{NeurIPS}, 33:\penalty0 16199--16210, 2020.

\bibitem[Kireev et~al.(2021)Kireev, Andriushchenko, and
  Flammarion]{advTrain_cprobust_arxiv21}
Klim Kireev, Maksym Andriushchenko, and Nicolas Flammarion.
\newblock On the effectiveness of adversarial training against common
  corruptions.
\newblock \emph{arXiv}, 2021.

\bibitem[Krizhevsky et~al.(2009)Krizhevsky, Hinton, et~al.]{db_cifar10}
Alex Krizhevsky, Geoffrey Hinton, et~al.
\newblock Learning multiple layers of features from tiny images.
\newblock 2009.

\bibitem[Kurakin et~al.(2016)Kurakin, Goodfellow, and
  Bengio]{advTrain_arxiv_2017}
Alexey Kurakin, Ian Goodfellow, and Samy Bengio.
\newblock Adversarial machine learning at scale.
\newblock \emph{arXiv preprint arXiv:1611.01236}, 2016.

\bibitem[Lecuyer et~al.(2019)Lecuyer, Atlidakis, Geambasu, Hsu, and
  Jana]{randSmoothDP_sp_2019}
Mathias Lecuyer, Vaggelis Atlidakis, Roxana Geambasu, Daniel Hsu, and Suman
  Jana.
\newblock Certified robustness to adversarial examples with differential
  privacy.
\newblock In \emph{IEEE SP}, 2019.

\bibitem[Lee et~al.(2019)Lee, Yuan, Chang, and Jaakkola]{certiLP_arxiv_2019}
Guang-He Lee, Yang Yuan, Shiyu Chang, and Tommi~S Jaakkola.
\newblock Tight certificates of adversarial robustness for randomly smoothed
  classifiers.
\newblock \emph{arXiv preprint arXiv:1906.04948}, 2019.

\bibitem[Li et~al.(2019)Li, Chen, Wang, and Carin]{certiSmoothing_nips_2019}
Bai Li, Changyou Chen, Wenlin Wang, and Lawrence Carin.
\newblock Certified adversarial robustness with additive noise.
\newblock \emph{NeurIPS}, 2019.

\bibitem[Li et~al.(2016)Li, Wang, Shi, Liu, and Hou]{li2016revisiting}
Yanghao Li, Naiyan Wang, Jianping Shi, Jiaying Liu, and Xiaodi Hou.
\newblock Revisiting batch normalization for practical domain adaptation.
\newblock \emph{arXiv}, 2016.

\bibitem[Liu et~al.(2018)Liu, Cheng, Zhang, and Hsieh]{smooth2_heu_2018}
Xuanqing Liu, Minhao Cheng, Huan Zhang, and Cho-Jui Hsieh.
\newblock Towards robust neural networks via random self-ensemble.
\newblock In \emph{ECCV}, 2018.

\bibitem[Madry et~al.(2018)Madry, Makelov, Schmidt, Tsipras, and
  Vladu]{madry_iclr_2018}
Aleksander Madry, Aleksandar Makelov, Ludwig Schmidt, Dimitris Tsipras, and
  Adrian Vladu.
\newblock Towards deep learning models resistant to adversarial attacks.
\newblock In \emph{ICLR}, 2018.

\bibitem[Mao et~al.(2021)Mao, Chiquier, Wang, Yang, and
  Vondrick]{reversible_iccv_2021}
Chengzhi Mao, Mia Chiquier, Hao Wang, Junfeng Yang, and Carl Vondrick.
\newblock Adversarial attacks are reversible with natural supervision.
\newblock \emph{arXiv preprint arXiv:2103.14222}, 2021.

\bibitem[Mirman et~al.(2018)Mirman, Gehr, and Vechev]{certify6_icml_2018}
Matthew Mirman, Timon Gehr, and Martin Vechev.
\newblock Differentiable abstract interpretation for provably robust neural
  networks.
\newblock In \emph{ICML}, 2018.

\bibitem[Moosavi~Dezfooli et~al.(2019)Moosavi~Dezfooli, Fawzi, Uesato, and
  Frossard]{cure_cvpr_2019}
Seyed~Mohsen Moosavi~Dezfooli, Alhussein Fawzi, Jonathan Uesato, and Pascal
  Frossard.
\newblock Robustness via curvature regularization, and vice versa.
\newblock In \emph{CVPR}, 2019.

\bibitem[Mueller et~al.(2021)Mueller, Balunovic, and
  Vechev]{certifyCombiner_iclr_2021}
Mark~Niklas Mueller, Mislav Balunovic, and Martin Vechev.
\newblock Certify or predict: Boosting certified robustness with compositional
  architectures.
\newblock In \emph{ICLR}, 2021.

\bibitem[Nado et~al.(2020)Nado, Padhy, Sculley, D'Amour, Lakshminarayanan, and
  Snoek]{adaptBN_cp_arxiv_2020}
Zachary Nado, Shreyas Padhy, D~Sculley, Alexander D'Amour, Balaji
  Lakshminarayanan, and Jasper Snoek.
\newblock Evaluating prediction-time batch normalization for robustness under
  covariate shift.
\newblock \emph{arXiv}, 2020.

\bibitem[Nandy et~al.(2020)Nandy, Hsu, and Lee]{rbfcnn_ijcnn_2020}
Jay Nandy, Wynne Hsu, and Mong{-}Li Lee.
\newblock Approximate manifold defense against multiple adversarial
  perturbations.
\newblock In \emph{IJCNN}, 2020.

\bibitem[Pang et~al.(2021)Pang, Yang, Dong, Su, and Zhu]{advTricks_iclr_2021}
Tianyu Pang, Xiao Yang, Yinpeng Dong, Hang Su, and Jun Zhu.
\newblock Bag of tricks for adversarial training.
\newblock In \emph{ICLR}, 2021.
\newblock URL \url{https://openreview.net/forum?id=Xb8xvrtB8Ce}.

\bibitem[Qin et~al.(2019)]{llrAdv_nips_2019}
Chongli Qin et~al.
\newblock Adversarial robustness through local linearization.
\newblock In \emph{NeurIPS}, 2019.

\bibitem[Raghunathan et~al.(2018)Raghunathan, Steinhardt, and
  Liang]{certify4_iclr_2018}
Aditi Raghunathan, Jacob Steinhardt, and Percy Liang.
\newblock Certified defenses against adversarial examples.
\newblock \emph{ICLR}, 2018.

\bibitem[Rice et~al.(2020)Rice, Wong, and Kolter]{advOverfitting_icml_2020}
Leslie Rice, Eric Wong, and Zico Kolter.
\newblock Overfitting in adversarially robust deep learning.
\newblock In \emph{ICML}, 2020.

\bibitem[Roy et~al.(2019)Roy, Siarohin, Sangineto, Bulo, Sebe, and
  Ricci]{roy2019unsupervised}
Subhankar Roy, Aliaksandr Siarohin, Enver Sangineto, Samuel~Rota Bulo, Nicu
  Sebe, and Elisa Ricci.
\newblock Unsupervised domain adaptation using feature-whitening and consensus
  loss.
\newblock In \emph{CVPR}, 2019.

\bibitem[Salman et~al.(2019{\natexlab{a}})Salman, Li, Razenshteyn, Zhang,
  Zhang, Bubeck, and Yang]{advSmooth_nips_2019}
Hadi Salman, Jerry Li, Ilya Razenshteyn, Pengchuan Zhang, Huan Zhang, Sebastien
  Bubeck, and Greg Yang.
\newblock Provably robust deep learning via adversarially trained smoothed
  classifiers.
\newblock In \emph{NeurIPS}, 2019{\natexlab{a}}.

\bibitem[Salman et~al.(2019{\natexlab{b}})Salman, Yang, Zhang, Hsieh, and
  Zhang]{certify9_nips_2019}
Hadi Salman, Greg Yang, Huan Zhang, Cho-Jui Hsieh, and Pengchuan Zhang.
\newblock A convex relaxation barrier to tight robustness verification of
  neural networks.
\newblock In \emph{NeurIPS}, 2019{\natexlab{b}}.

\bibitem[Salman et~al.(2020)Salman, Sun, Yang, Kapoor, and
  Kolter]{advDenoised_nips_2020}
Hadi Salman, Mingjie Sun, Greg Yang, Ashish Kapoor, and J~Zico Kolter.
\newblock Denoised smoothing: A provable defense for pretrained classifiers.
\newblock \emph{NeurIPS}, 2020.

\bibitem[Schneider et~al.(2020)Schneider, Rusak, Eck, Bringmann, Brendel, and
  Bethge]{adaptBN_cp_nips_2020}
Steffen Schneider, Evgenia Rusak, Luisa Eck, Oliver Bringmann, Wieland Brendel,
  and Matthias Bethge.
\newblock Improving robustness against common corruptions by covariate shift
  adaptation.
\newblock \emph{NeurIPS}, 2020.

\bibitem[Schott et~al.(2019)Schott, Rauber, Bethge, and
  Brendel]{mnist_multiple_iclr2019}
Lukas Schott, Jonas Rauber, Matthias Bethge, and Wieland Brendel.
\newblock Towards the first adversarially robust neural network model on
  {MNIST}.
\newblock In \emph{ICLR}, 2019.

\bibitem[Sheikholeslami et~al.(2021)Sheikholeslami, Lotfi, and
  Kolter]{certifyAbstain_iclr_2021}
Fatemeh Sheikholeslami, Ali Lotfi, and J~Zico Kolter.
\newblock Provably robust classification of adversarial examples with
  detection.
\newblock In \emph{ICLR}, 2021.

\bibitem[Sun et~al.(2017)Sun, Feng, and Saenko]{da1_2017}
Baochen Sun, Jiashi Feng, and Kate Saenko.
\newblock Correlation alignment for unsupervised domain adaptation.
\newblock In \emph{Domain Adaptation in Computer Vision Applications}, pp.\
  153--171. Springer, 2017.

\bibitem[Sun et~al.(2020)Sun, Wang, Liu, Miller, Efros, and
  Hardt]{adapt_cp_icml_2020}
Yu~Sun, Xiaolong Wang, Zhuang Liu, John Miller, Alexei Efros, and Moritz Hardt.
\newblock Test-time training with self-supervision for generalization under
  distribution shifts.
\newblock In \emph{ICML}, 2020.

\bibitem[Szegedy et~al.(2014)]{advStart_iclr_2014}
Christian Szegedy et~al.
\newblock Intriguing properties of neural networks.
\newblock In \emph{ICLR}, 2014.

\bibitem[Tjeng et~al.(2017)Tjeng, Xiao, and Tedrake]{exact1_iclr_2017}
Vincent Tjeng, Kai Xiao, and Russ Tedrake.
\newblock Evaluating robustness of neural networks with mixed integer
  programming.
\newblock \emph{ICLR}, 2017.

\bibitem[Tram{\`e}r \& Boneh(2019)Tram{\`e}r and Boneh]{multiple_advTrain_2019}
Florian Tram{\`e}r and Dan Boneh.
\newblock Adversarial training and robustness for multiple perturbations.
\newblock In \emph{NeurIPS}, 2019.

\bibitem[Tsipras et~al.(2019)Tsipras, Santurkar, Engstrom, Turner, and
  Madry]{robustnessDrops_iclr_2018}
Dimitris Tsipras, Shibani Santurkar, Logan Engstrom, Alexander Turner, and
  Aleksander Madry.
\newblock Robustness may be at odds with accuracy.
\newblock In \emph{ICLR}, 2019.

\bibitem[Uesato et~al.(2018)Uesato, O'Donoghue, Oord, and
  Kohli]{spsa_icml_2018}
Jonathan Uesato, Brendan O'Donoghue, Aaron van~den Oord, and Pushmeet Kohli.
\newblock Adversarial risk and the dangers of evaluating against weak attacks.
\newblock \emph{ICML}, 2018.

\bibitem[Uesato et~al.(2019)Uesato, Alayrac, Huang, Stanforth, Fawzi, and
  Kohli]{advUnlabelled2_arxiv_2019}
Jonathan Uesato, Jean-Baptiste Alayrac, Po-Sen Huang, Robert Stanforth,
  Alhussein Fawzi, and Pushmeet Kohli.
\newblock Are labels required for improving adversarial robustness?
\newblock \emph{arXiv}, 2019.

\bibitem[Wang et~al.(2020{\natexlab{a}})Wang, Shelhamer, Liu, Olshausen, and
  Darrell]{adaptation4_2020}
Dequan Wang, Evan Shelhamer, Shaoteng Liu, Bruno Olshausen, and Trevor Darrell.
\newblock Tent: Fully test-time adaptation by entropy minimization.
\newblock \emph{arXiv}, 2020{\natexlab{a}}.

\bibitem[Wang et~al.(2020{\natexlab{b}})Wang, Chen, Gui, Hu, Liu, and
  Wang]{dualBN_2}
Haotao Wang, Tianlong Chen, Shupeng Gui, TingKuei Hu, Ji~Liu, and Zhangyang
  Wang.
\newblock Once-for-all adversarial training: In-situ tradeoff between
  robustness and accuracy for free.
\newblock \emph{NeurIPS}, 33:\penalty0 7449--7461, 2020{\natexlab{b}}.

\bibitem[Wang et~al.(2021)Wang, Xiao, Kossaifi, Yu, Anandkumar, and
  Wang]{dualBN_3}
Haotao Wang, Chaowei Xiao, Jean Kossaifi, Zhiding Yu, Anima Anandkumar, and
  Zhangyang Wang.
\newblock Augmax: Adversarial composition of random augmentations for robust
  training.
\newblock \emph{NeurIPS}, 34:\penalty0 237--250, 2021.

\bibitem[Wang et~al.(2018)Wang, Pei, Whitehouse, Yang, and
  Jana]{certify3_nips_2018}
Shiqi Wang, Kexin Pei, Justin Whitehouse, Junfeng Yang, and Suman Jana.
\newblock Efficient formal safety analysis of neural networks.
\newblock In \emph{NeurIPS}, 2018.

\bibitem[Weng et~al.(2018)Weng, Zhang, Chen, Song, Hsieh, Daniel, Boning, and
  Dhillon]{certify14_icml_2018}
Lily Weng, Huan Zhang, Hongge Chen, Zhao Song, Cho-Jui Hsieh, Luca Daniel,
  Duane Boning, and Inderjit Dhillon.
\newblock Towards fast computation of certified robustness for relu networks.
\newblock In \emph{ICML}, 2018.

\bibitem[Wong \& Kolter(2018)Wong and Kolter]{certify1_icml_2018}
Eric Wong and Zico Kolter.
\newblock Provable defenses against adversarial examples via the convex outer
  adversarial polytope.
\newblock In \emph{ICML}, 2018.

\bibitem[Wong et~al.(2018)Wong, Schmidt, Metzen, and
  Kolter]{certify8_nips_2018}
Eric Wong, Frank~R Schmidt, Jan~Hendrik Metzen, and J~Zico Kolter.
\newblock Scaling provable adversarial defenses.
\newblock \emph{NeurIPS}, 2018.

\bibitem[Wong et~al.(2020)Wong, Rice, and Kolter]{fgsmAdvTrain_iclr_2020}
Eric Wong, Leslie Rice, and J~Zico Kolter.
\newblock Fast is better than free: Revisiting adversarial training.
\newblock \emph{ICLR}, 2020.

\bibitem[Xie \& Yuille(2020)Xie and Yuille]{mixtureBN_iclr2020}
Cihang Xie and Alan Yuille.
\newblock Intriguing properties of adversarial training at scale.
\newblock \emph{ICLR}, 2020.

\bibitem[Xie et~al.(2020{\natexlab{a}})Xie, Tan, Gong, Wang, Yuille, and
  Le]{auxBN_cvpr_2020}
Cihang Xie, Mingxing Tan, Boqing Gong, Jiang Wang, Alan~L Yuille, and Quoc~V
  Le.
\newblock Adversarial examples improve image recognition.
\newblock In \emph{IEEE CVPR}, 2020{\natexlab{a}}.

\bibitem[Xie et~al.(2020{\natexlab{b}})Xie, Luong, Hovy, and
  Le]{adaptation3_cvpr_2020}
Qizhe Xie, Minh-Thang Luong, Eduard Hovy, and Quoc~V Le.
\newblock Self-training with noisy student improves imagenet classification.
\newblock In \emph{CVPR}, 2020{\natexlab{b}}.

\bibitem[Yang et~al.(2020)Yang, Duan, Hu, Salman, Razenshteyn, and
  Li]{anyLpCertify_arxiv_2020}
Greg Yang, Tony Duan, Edward Hu, Hadi Salman, Ilya Razenshteyn, and Jerry Li.
\newblock Randomized smoothing of all shapes and sizes.
\newblock \emph{arXiv}, 2020.

\bibitem[Zhai et~al.(2020)Zhai, Dan, He, Zhang, Gong, Ravikumar, Hsieh, and
  Wang]{macer_iclr_2020}
Runtian Zhai, Chen Dan, Di~He, Huan Zhang, Boqing Gong, Pradeep Ravikumar,
  Cho-Jui Hsieh, and Liwei Wang.
\newblock Macer: Attack-free and scalable robust training via maximizing
  certified radius.
\newblock In \emph{ICLR}, 2020.

\bibitem[Zhang et~al.(2019)]{treadAdv_icml_2019}
Hongyang Zhang et~al.
\newblock Theoretically principled trade-off between robustness and accuracy.
\newblock In \emph{ICML}, 2019.

\bibitem[Zhang et~al.(2020)Zhang, Xu, Han, Niu, Cui, Sugiyama, and
  Kankanhalli]{fat_icml_2020}
Jingfeng Zhang, Xilie Xu, Bo~Han, Gang Niu, Lizhen Cui, Masashi Sugiyama, and
  Mohan Kankanhalli.
\newblock Attacks which do not kill training make adversarial learning
  stronger.
\newblock In \emph{ICML}. PMLR, 2020.

\end{thebibliography}
\bibliographystyle{tmlr}

\newpage
\appendix
\section{Appendix}

\subsection{Detailed Certification results for different models using proposed method}

\begin{table}[h]
\centering
\resizebox{15.0cm}{!} 
{%
\begin{tabular}{ll!{\vrule width 1.5pt}ccccccccccccc!{\vrule width 1.5pt}c}
\Xhline{3\arrayrulewidth}
\multicolumn{15}{c}{{ImageNet}}
\\
\Xhline{3\arrayrulewidth}
Model & Noise-level & \multicolumn{8}{c}{$\ell_2$ Radius } \\
&& 0.0 & 0.25 & 0.5 & 0.75 & 1.0 & 1.25 & 1.5 & 1.75 & 2.0 & 2.25 & 2.5 &2.75 & 3.0 & ACR
\\ 
\hline 
\\
Baseline & $\sigma = $ 0.25
& 13.6 & 7.8 & 4.8 & 3.0 & 0.0 & 0.0 & 0.0 & 0.0 & 0.0 & 0.0 & 0.0 & 0.0 & 0.0 & 0.055
\\ 
\rowcolor{Gray} \cellcolor{Gray}
Baseline & Auto-Noise
& 52.4 & 36.8 & 5.0 & 3.2 & 0.6 & 0.6 & 0.4 & 0.4 & 0.4 & 0.4 & 0.4 & 0.4 & 0.4 & 0.194
\\ 
\rowcolor{Gray} \cellcolor{Gray}
Baseline + Adaptation & Auto-Noise
& 57.4 & 25.8 & 4.6 & 1.8 & 0.4 & 0.2 & 0.0 & 0.0 & 0.0 & 0.0 & 0.0 & 0.0 & 0.0 & 0.15
\\ 
\Xhline{2\arrayrulewidth} \\

\rowcolor{Gray} \cellcolor{Gray}
Adv$_{\infty}[4/255]$ & Auto-Noise
& 34.6 & 31.2 & 4.0 & 3.8 & 0.6 & 0.6 & 0.4 & 0.4 & 0.2 & 0.2 & 0.0 & 0.0 & 0.0 & 0.169
\\ 
\rowcolor{Gray} \cellcolor{Gray}
Adv$_{\infty}[4/255]$ + Adaptation & $\sigma = $ 0.5
& 47.4 & 43.6 & 39.4 & 35.8 & 31.4 & 27.6 & 23.4 & 18.2 & 0.0 & 0.0 & 0.0 & 0.0 & 0.0 & 0.609
\\
\rowcolor{Gray} \cellcolor{Gray}
Adv$_{\infty}[4/255]$ + Adaptation & Auto-Noise
& 65.6 & 59.4 & 50.6 & 46.6 & 38.0 & 33.2 & 26.2 & 20.8 & 12.0 & 8.6 & 6.0 & 4.0 & 1.6 & 0.859
\\ 
\Xhline{2\arrayrulewidth} \\

\rowcolor{Gray} \cellcolor{Gray}
Adv$_{2}[3.0]$ & Auto-Noise
& 44.0 & 41.6 & 11.8 & 11.2 & 2.8 & 2.6 & 0.4 & 0.4 & 0.0 & 0.0 & 0.0 & 0.0 & 0.0 & 0.261
\\ 
\rowcolor{Gray} \cellcolor{Gray}
Adv$_{2}[3.0]$ + Adaptation & $\sigma = $ 0.5
& 50.2 & 47.0 & 43.0 & 39.0 & 36.4 & 32.8 & 30.8 & 27.0 & 0.0 & 0.0 & 0.0 & 0.0 & 0.0 & 0.711
\\
\rowcolor{Gray} \cellcolor{Gray}
Adv$_{2}[3.0]$ + Adaptation & Auto-Noise
& 66.6 & 63.8 & 58.6 & 55.4 & 45.6 & \textbf{41.0} & \textbf{35.8} & \textbf{32.4} & \textbf{23.6} & \textbf{18.6} & \textbf{15.0} & \textbf{12.8} & \textbf{7.4} & \textbf{1.148}
\\ 
\Xhline{2\arrayrulewidth} \\

Rand$_{\sigma=0.5}$ & $\sigma = $ 0.50
& 68.2 & 60.8 & 54.4 & 47.8 & 38.8 & 33.8 & 28.6 & 23.4 & 0.0 & 0.0 & 0.0 & 0.0 & 0.0 & 0.811
\\ 
\rowcolor{Gray} \cellcolor{Gray}
Rand$_{\sigma=0.5}$ & Auto-Noise
& 71.4 & 65.6 & 58.8 & 51.4 & 42.8 & 37.0 & 28.8 & 23.8 & 2.6 & 1.6 & 0.2 & 0.2 & 0.2 & 0.88
\\ 
\rowcolor{Gray} \cellcolor{Gray}
Rand$_{\sigma=0.5}$ + Adaptation & Auto-Noise
& \textbf{74.8} & \textbf{69.8} & \textbf{64.4} & \textbf{56.6} & \textbf{47.8} & 40.0 & 34.4 & 27.4 & 20.2 & 15.8 & 10.4 & 6.4 & 3.2 & 1.095
\\ 
\Xhline{2\arrayrulewidth}
\end{tabular}%
}
\caption{
{ImageNet}: Certified top-1 accuracy at various $\ell_2$ radii as we vary $\sigma$ for BN adaptation and certification along with average certified radii (ACR).
We use ResNet50 for {ImageNet}.}
\label{table_app_imagenet_certify}
\end{table}

\begin{table}[!h]
\centering
\resizebox{16.5cm}{!} 
{%
\begin{tabular}{lc!{\vrule width 1.5pt}ccccccccccccc!{\vrule width 1.5pt}c}
\Xhline{3\arrayrulewidth}
\multicolumn{15}{c}{{ImageNet}}
\\
\Xhline{3\arrayrulewidth}
Model & Noise-level & \multicolumn{8}{c}{$\ell_2$ Radius } \\
&& 0.0 & 0.25 & 0.5 & 0.75 & 1.0 & 1.25 & 1.5 & 1.75 & 2.0 & 2.25 & 2.5 &2.75 & 3.0 & ACR
\\ 
\hline 
\\
Baseline & $\sigma = $ 0.25
& 10.49 & 6.96 & 2.04 & 0.09 & 0.0 & 0.0 & 0.0 & 0.0 & 0.0 & 0.0 & 0.0 & 0.0 & 0.0 & 0.035
\\
\rowcolor{Gray} \cellcolor{Gray}
Baseline & Auto-Noise
& 33.57 & 18.56 & 10.25 & 4.44 & 0.83 & 0.07 & 0.01 & 0.0 & 0.0 & 0.0 & 0.0 & 0.0 & 0.0 & 0.124
\\
\rowcolor{Gray} \cellcolor{Gray}
Baseline + Adaptation & Auto-Noise
& 59.64 & 21.66 & 7.81 & 3.97 & 1.28 & 0.36 & 0.07 & 0.0 & 0.0 & 0.0 & 0.0 & 0.0 & 0.0 & 0.154
\\ \hline
\\

\rowcolor{Gray} \cellcolor{Gray}
Adv$_{\infty}[4/255]$ & Auto-Noise
& 73.71 & 63.04 & 28.75 & 23.91 & 17.96 & 14.19 & 10.12 & 7.08 & 4.99 & 4.14 & 3.24 & 2.49 & 1.84 & 0.57
\\
\rowcolor{Gray} \cellcolor{Gray}
Adv$_{\infty}[4/255]$ + Adaptation & $\sigma = 0.5$
& 63.23 & 47.34 & 31.83 & 18.78 & 9.98 & 4.44 & 1.62 & 0.28 & 0.0 & 0.0 & 0.0 & 0.0 & 0.0 & 0.361
\\
\rowcolor{Gray} \cellcolor{Gray}
Adv$_{\infty}[4/255]$ + Adaptation & Auto-Noise
& 85.53 & 76.02 & 49.42 & 36.23 & 21.05 & 13.4 & 8.7 & 5.58 & 3.15 & 1.69 & 0.66 & 0.19 & 0.07 & 0.654
\\ \hline
\rowcolor{Gray} \cellcolor{Gray}
Adv$_{\infty}[8/255]$ & Auto-Noise
& 74.46 & 66.57 & 36.32 & 30.15 & 18.87 & 13.71 & 8.7 & 5.29 & 2.89 & 1.88 & 1.17 & 0.75 & 0.46 & 0.578
\\
\rowcolor{Gray} \cellcolor{Gray}
Adv$_{\infty}[8/255]$ + Adaptation & $\sigma = 0.5$
& 64.2 & 53.65 & 42.91 & 32.58 & 22.68 & 14.24 & 7.88 & 2.94 & 0.0 & 0.0 & 0.0 & 0.0 & 0.0 & 0.52
\\
\rowcolor{Gray} \cellcolor{Gray}
Adv$_{\infty}[8/255]$ + Adaptation & Auto-Noise
& 79.69 & 71.78 & 53.76 & 43.61 & 29.68 & 20.87 & 14.04 & 9.08 & 5.53 & 3.33 & 1.71 & 0.81 & 0.35 & 0.741
\\ \hline
\rowcolor{Gray} \cellcolor{Gray}
Adv$_{\infty}[12/255]$ & Auto-Noise
& 69.46 & 63.12 & 35.73 & 30.63 & 17.54 & 14.78 & 10.27 & 9.16 & 8.01 & 7.2 & 6.21 & 5.31 & 3.78 & 0.649
\\ 
\rowcolor{Gray} \cellcolor{Gray}
Adv$_{\infty}[12/255]$ + Adaptation & $\sigma = 0.5$
& 59.19 & 51.53 & 43.94 & 36.41 & 28.69 & 21.25 & 14.53 & 8.03 & 0.0 & 0.0 & 0.0 & 0.0 & 0.0 & 0.583
\\
\rowcolor{Gray} \cellcolor{Gray}
Adv$_{\infty}[12/255]$ + Adaptation & Auto-Noise
& 70.75 & 64.54 & 50.63 & 43.38 & 32.5 & 24.43 & 18.05 & 12.2 & 8.31 & 5.37 & 3.36 & 1.96 & 1.23 & 0.76
\\ \hline
\rowcolor{Gray} \cellcolor{Gray}
Adv$_{\infty}[16/255]$ & Auto-Noise
& 58.47 & 53.41 & 31.48 & 26.76 & 16.68 & 14.04 & 10.89 & 9.17 & 7.42 & 5.62 & 3.81 & 2.43 & 1.53 & 0.55
\\
\rowcolor{Gray} \cellcolor{Gray}
Adv$_{\infty}[16/255]$ + Adaptation & $\sigma = 0.5$
& 53.8 & 48.07 & 42.51 & 36.54 & 30.55 & 24.68 & 18.49 & 12.11 & 0.0 & 0.0 & 0.0 & 0.0 & 0.0 & 0.599
\\
\rowcolor{Gray} \cellcolor{Gray}
Adv$_{\infty}[16/255]$ + Adaptation & Auto-Noise
& 61.07 & 56.01 & 45.76 & 40.29 & 31.6 & 24.53 & 18.73 & 14.23 & 10.39 & 7.37 & 5.05 & 3.36 & 2.13 & 0.731
\\ \hline
\\
\rowcolor{Gray} \cellcolor{Gray}
Adv$_{2}[0.5]$ & Auto-Noise
& 71.6 & 61.2 & 22.17 & 17.64 & 12.26 & 10.97 & 10.17 & 9.76 & 9.48 & 9.12 & 8.53 & 7.52 & 6.61 & 0.603
\\ 
\rowcolor{Gray} \cellcolor{Gray}
Adv$_{2}[0.5]$ + Adaptation & $\sigma = 0.5$
& 63.77 & 48.81 & 33.82 & 20.95 & 11.5 & 5.64 & 2.29 & 0.62 & 0.0 & 0.0 & 0.0 & 0.0 & 0.0 & 0.386
\\
\rowcolor{Gray} \cellcolor{Gray}
Adv$_{2}[0.5]$ + Adaptation & Auto-Noise
& 86.26 & 77.52 & 61.25 & 46.44 & 23.42 & 16.22 & 11.55 & 9.2 & 7.62 & 6.47 & 5.07 & 3.81 & 2.51 & 0.796
\\ \hline
\rowcolor{Gray} \cellcolor{Gray}
Adv$_{2}[1.0]$ & Auto-Noise
& 81.08 & 74.0 & 43.82 & 34.77 & 17.25 & 11.15 & 6.19 & 3.68 & 1.76 & 1.02 & 0.52 & 0.27 & 0.15 & 0.608
\\ 
\rowcolor{Gray} \cellcolor{Gray}
Adv$_{2}[1.0]$ + Adaptation & $\sigma = 0.5$
& 66.05 & 56.45 & 46.24 & 35.6 & 26.89 & 18.73 & 11.37 & 5.41 & 0.0 & 0.0 & 0.0 & 0.0 & 0.0 & 0.582
\\
\rowcolor{Gray} \cellcolor{Gray}
Adv$_{2}[1.0]$ + Adaptation & Auto-Noise
& 82.34 & 75.38 & 64.53 & 53.8 & 36.04 & 27.55 & 19.53 & 12.61 & 7.29 & 4.33 & 2.39 & 1.21 & 0.67 & 0.871
\\ \hline
\rowcolor{Gray} \cellcolor{Gray}
Adv$_{2}[1.25]$ & Auto-Noise
& 78.32 & 72.56 & 44.86 & 39.13 & 25.98 & 21.02 & 16.31 & 12.17 & 8.18 & 5.18 & 2.25 & 1.25 & 0.66 & 0.741
\\ 
\rowcolor{Gray} \cellcolor{Gray}
Adv$_{2}[1.25]$ + Adaptation & $\sigma = 0.5$
& 66.22 & 57.73 & 48.8 & 39.64 & 31.07 & 22.61 & 15.82 & 8.96 & 0.0 & 0.0 & 0.0 & 0.0 & 0.0 & 0.644
\\
\rowcolor{Gray} \cellcolor{Gray}
Adv$_{2}[1.25]$ + Adaptation & Auto-Noise
& 80.52 & 74.25 & 64.95 & 56.56 & 40.81 & 32.71 & 24.96 & 17.87 & 11.5 & 8.07 & 5.88 & 4.05 & 2.59 & 0.972
\\ \hline
\rowcolor{Gray} \cellcolor{Gray}
Adv$_{2}[1.5]$ & Auto-Noise
& 75.26 & 69.75 & 48.42 & 41.73 & 26.44 & 20.82 & 14.12 & 10.14 & 6.87 & 4.75 & 3.02 & 1.81 & 0.87 & 0.733
\\ 
\rowcolor{Gray} \cellcolor{Gray}
Adv$_{2}[1.5]$ + Adaptation & $\sigma = 0.5$
& 63.67 & 56.55 & 49.19 & 41.72 & 34.47 & 27.36 & 20.23 & 12.98 & 0.0 & 0.0 & 0.0 & 0.0 & 0.0 & 0.687
\\
\rowcolor{Gray} \cellcolor{Gray}
Adv$_{2}[1.5]$ + Adaptation & Auto-Noise
& 76.22 & 70.47 & 62.39 & 55.91 & 42.45 & 35.79 & 29.01 & 21.71 & 14.23 & 10.04 & 6.51 & 4.07 & 2.29 & 0.99
\\ \hline
\rowcolor{Gray} \cellcolor{Gray}
Adv$_{2}[2.5]$ & Auto-Noise
& 61.2 & 57.42 & 42.26 & 38.0 & 26.56 & 21.41 & 14.21 & 10.34 & 7.33 & 4.96 & 3.24 & 2.06 & 1.07 & 0.662
\\ 
\rowcolor{Gray} \cellcolor{Gray}
Adv$_{2}[2.5]$ + Adaptation & $\sigma = 0.5$
& 54.73 & 50.53 & 46.26 & 41.84 & 37.74 & 33.2 & 28.69 & 23.34 & 0.0 & 0.0 & 0.0 & 0.0 & 0.0 & 0.726
\\
\rowcolor{Gray} \cellcolor{Gray}
Adv$_{2}[2.5]$ + Adaptation & Auto-Noise
& 63.36 & 59.53 & 54.68 & 50.3 & 42.95 & 38.62 & 33.97 & 29.42 & 23.03 & 19.22 & 15.3 & 11.34 & 7.51 & 1.073
\\ \hline
\rowcolor{Gray} \cellcolor{Gray}
Adv$_{2}[3.0]$ & Auto-Noise
& 64.45 & 60.57 & 45.73 & 41.06 & 28.48 & 22.92 & 15.1 & 10.77 & 7.23 & 4.8 & 2.77 & 1.67 & 1.1 & 0.702
\\ 
\rowcolor{Gray} \cellcolor{Gray}
Adv$_{2}[3.0]$ + Adaptation & $\sigma = 0.5$
& 53.75 & 49.41 & 45.57 & 41.52 & 37.43 & 33.37 & 28.82 & 23.65 & 0.0 & 0.0 & 0.0 & 0.0 & 0.0 & 0.721
\\
\rowcolor{Gray} \cellcolor{Gray}
Adv$_{2}[3.0]$ + Adaptation & Auto-Noise
& 61.96 & 58.58 & 53.64 & 49.67 & 42.76 & 38.69 & 34.54 & 30.36 & 24.65 & 20.77 & 17.09 & 13.66 & 9.18 & 1.102
\\ \hline
\\
Rand$_{\sigma=0.5}$ & $\sigma = 0.5$
& 62.13 & 51.68 & 40.38 & 30.25 & 20.81 & 13.36 & 7.71 & 3.38 & 0.0 & 0.0 & 0.0 & 0.0 & 0.0 & 0.494
\\ 
\rowcolor{Gray} \cellcolor{Gray}
Rand$_{\sigma=0.5}$ & Auto-Noise
& 79.48 & 71.75 & 60.23 & 48.72 & 35.97 & 25.16 & 15.09 & 10.13 & 6.98 & 5.58 & 4.18 & 2.94 & 1.76 & 0.821
\\
\rowcolor{Gray} \cellcolor{Gray}
Rand$_{\sigma=0.5}$ + Adaptation & Auto-Noise
& 78.27 & 69.51 & 57.16 & 44.73 & 31.19 & 19.27 & 10.4 & 4.25 & 1.65 & 0.57 & 0.14 & 0.01 & 0.0 & 0.695
\\ \hline
\\
SmoothAdv$_{\sigma=0.5, \epsilon=0.25}$ & $\sigma = 0.5$
& 67.35 & 57.8 & 47.63 & 37.41 & 27.88 & 20.33 & 13.53 & 8.03 & 0.0 & 0.0 & 0.0 & 0.0 & 0.0 & 0.614
\\
\rowcolor{Gray} \cellcolor{Gray}
SmoothAdv$_{\sigma=0.5, \epsilon=0.25}$ & Auto-Noise
& 72.88 & 65.23 & 55.25 & 44.87 & 34.93 & 24.79 & 16.44 & 9.17 & 4.92 & 1.91 & 0.73 & 0.21 & 0.07 & 0.737
\\
\rowcolor{Gray} \cellcolor{Gray}
SmoothAdv$_{\sigma=0.5, \epsilon=0.25}$ + Adaptation & Auto-Noise
& 74.45 & 65.68 & 54.28 & 42.59 & 31.76 & 21.61 & 13.19 & 7.16 & 3.41 & 1.43 & 0.52 & 0.16 & 0.05 & 0.697
\\ \hline
SmoothAdv$_{\sigma=0.5, \epsilon=0.5}$ & $\sigma = 0.5$
& 67.21 & 58.82 & 49.68 & 40.35 & 31.93 & 24.18 & 17.05 & 10.57 & 0.0 & 0.0 & 0.0 & 0.0 & 0.0 & 0.665
\\
\rowcolor{Gray} \cellcolor{Gray}
SmoothAdv$_{\sigma=0.5, \epsilon=0.5}$ & Auto-Noise
& 71.85 & 65.28 & 56.41 & 48.2 & 39.47 & 29.82 & 21.44 & 13.55 & 7.54 & 3.45 & 1.35 & 0.41 & 0.12 & 0.807
\\
\rowcolor{Gray} \cellcolor{Gray}
SmoothAdv$_{\sigma=0.5, \epsilon=0.5}$ + Adaptation & Auto-Noise
& 73.54 & 66.14 & 56.81 & 47.04 & 37.2 & 26.99 & 18.46 & 10.78 & 5.67 & 2.6 & 1.0 & 0.44 & 0.13 & 0.773
\\ \hline
SmoothAdv$_{\sigma=0.5, \epsilon=1.0}$ & $\sigma = 0.5$
& 63.95 & 56.53 & 49.53 & 41.38 & 34.63 & 27.81 & 21.22 & 14.41 & 0.0 & 0.0 & 0.0 & 0.0 & 0.0 & 0.694
\\
\rowcolor{Gray} \cellcolor{Gray}
SmoothAdv$_{\sigma=0.5, \epsilon=1.0}$ & Auto-Noise
& 67.91 & 62.28 & 55.36 & 48.41 & 41.14 & 33.57 & 26.27 & 18.42 & 11.87 & 6.1 & 2.64 & 0.95 & 0.32 & 0.854
\\
\rowcolor{Gray} \cellcolor{Gray}
SmoothAdv$_{\sigma=0.5, \epsilon=1.0}$ + Adaptation & Auto-Noise
& 69.6 & 63.04 & 55.52 & 47.65 & 38.93 & 30.7 & 22.57 & 15.34 & 9.21 & 4.51 & 2.02 & 0.9 & 0.33 & 0.813
\\ \hline
SmoothAdv$_{\sigma=0.5, \epsilon=2.0}$ & $\sigma = 0.5$
& 57.59 & 52.82 & 47.67 & 42.68 & 37.55 & 32.64 & 27.52 & 22.42 & 0.0 & 0.0 & 0.0 & 0.0 & 0.0 & 0.733
\\
\rowcolor{Gray} \cellcolor{Gray}
SmoothAdv$_{\sigma=0.5, \epsilon=2.0}$ & Auto-Noise
& 61.27 & 57.27 & 52.52 & 48.17 & 43.49 & 38.02 & 33.15 & 27.47 & 21.86 & 15.81 & 9.5 & 4.97 & 2.01 & 0.965
\\
\rowcolor{Gray} \cellcolor{Gray}
SmoothAdv$_{\sigma=0.5, \epsilon=2.0}$ + Adaptation & Auto-Noise
& 61.23 & 56.9 & 51.33 & 46.44 & 41.05 & 35.65 & 30.11 & 24.35 & 18.48 & 12.8 & 7.76 & 4.27 & 2.3 & 0.908
\\ \hline
\Xhline{3\arrayrulewidth}
\end{tabular}%
}
\vspace*{-0.5em}
\caption{
{CIFAR-10}: Certified top-1 accuracy at various $\ell_2$ radii as we vary $\sigma$ for test-time BN adaptation along with average certified radii (ACR) for individual settings.
}
\label{table_app_cifar10_certify}
\end{table}

\newpage
\subsection{Implementation Details}
\label{sec_implementation}
We present our experimental results on {CIFAR-10} \citep{db_cifar10} and {ImageNet} \citep{db_imagenet} datasets.
The descriptions of different models and training hyper-parameters are provided in the following:

\subsubsection{{CIFAR-10}. }
We use pre-activation ResNet18 architecture \citep{resnet_eccv_2016} for our experiments on {CIFAR-10}.
We apply the SGD optimizer with a batch size of $128$. 
We execute a total of $200$ training epochs and apply a step-wise learning rate decay set initially at $0.1$ and divided by $10$ at $100$ and $150$ epochs, and weight decay $5 \times 10^{-4}$.

\textbf{AT models \citep{madry_iclr_2018,advOverfitting_icml_2020}: }
Unless and otherwise specified, our AT models are learned using early stopping criteria as described in \citep{advOverfitting_icml_2020}.
We learn several AT models with different threat boundaries for our experiments.
We denote them by specifying their corresponding threat model and threat boundaries.
For example, Adv$_2[1.5]$ denotes an AT model that is learned using PGD adversary with $\ell_2$ threat model and a threat boundary of $\epsilon = 1.5$, along with \textit{early-stopping criteria} \citep{advOverfitting_icml_2020}.
We also learn AT models \textit{without} using early-stopping criteria, as in \citep{madry_iclr_2018} for our comparison in Figure \ref{fig:certify_overfit}. 
These models are denoted as Adv$^{overfit}$.

We use \textit{projected gradient descent (PGD)} adversarial attack \citep{madry_iclr_2018} to train these AT models as follows:
For Adv$_{\infty}$, we use $10$ iterations and an $\ell_{\infty}$ step size of $\epsilon/4$.
For Adv$_{2}$, we use $10$ iterations and an $\ell_{2}$ step size of $\epsilon/8.5$.
This is the same experimental setup as in \citep{advOverfitting_icml_2020}).
We choose a small set of $1,000$ images from the {CIFAR-10} test set for our validation. 
We apply the PGD attack with the same hyper-parameters for our validation during training. 
We save the best model using the \textit{early-stopping} criteria \citep{advOverfitting_icml_2020}.

\textbf{Randomized smoothing model by \cite{certiSmoothing_icml_2019}: }
We also train Rand$_{\sigma=0.5}$ by training with augmented random noise, sampled from an isotropic Gaussian distribution $\mathcal{N}(0,\sigma^2I)$ with $\sigma=0.5$.
Here, we keep the same model architecture, learning rates, batch sizes, and other hyper-parameters as used to learn the AT models.

\textbf{Randomized smoothing model by \cite{advSmooth_nips_2019}: }
We also compare with the state-of-the-art certification models, called `SmoothAdv', by \cite{advSmooth_nips_2019} for our experiments on $\ell_2$ certification
We train the SmoothAdv models by choosing random noise vectors followed by an adaptive adversarial attack with specified $\ell_2$ threat boundary of $\epsilon$ at each iteration.
The noise vectors are sampled from an isotropic Gaussian distribution $\mathcal{N}(0,\sigma^2I)$.

We note that the training hyper-parameter $\epsilon$ has the most significant impact on the certification curve for a SmoothAdv model (please refer to Table 7-15 of \citep{advSmooth_nips_2019} for more details).
For our experiments, we train $4$ different SmoothAdv models with $\epsilon = \{0.25, 0.5, 1, 2\}$ and $\sigma=0.5$ using adaptive PGD attack with $10$ steps.
We denote them as SmoothAdv$_{\sigma=0.5, \epsilon= 0.25}$, SmoothAdv$_{\sigma=0.5, \epsilon= 0.5}$, SmoothAdv$_{\sigma=0.5, \epsilon= 1}$ and SmoothAdv$_{\sigma=0.5, \epsilon= 2}$ respectively.
We use the same training set-up and other hyper-parameters as specified in their Github:  \href{https://github.com/Hadisalman/smoothing-adversarial}{https://github.com/Hadisalman/smoothing-adversarial}.

\subsubsection{{ImageNet}. }
We use ResNet50 architecture \citep{resnet_cvpr_2016} for {ImageNet}.
We obtain the Baseline and Rand$_{\sigma=0.5}$ models from \citep{certiSmoothing_icml_2019}\footnote{\href{https://github.com/locuslab/smoothing}{https://github.com/locuslab/smoothing}}.
These models are trained using Gaussian augmented noises, sampled from isotropic Gaussian distribution $\mathcal{N}(0,\sigma^2I)$ with $\sigma=0.0$ (i.e., no noise) and $\sigma=0.5$ respectively.

The AT models i.e., Adv$_{\infty}[4/255]$ and Adv$_{2}[3.0]$ are learned for $\ell_{\infty}$ and $\ell_2$ threat models with threat boundary of $4/255$ and $3$, respectively.
We use the publicly available models provided by \cite{advOverfitting_icml_2020}
\footnote{\href{https://github.com/locuslab/robust_overfitting}{https://github.com/locuslab/robust\_overfitting}}. 
These models are fine-tuned using PGD-based adversarial training with early stopping criteria, originally provided by \cite{advImageNetModels_2019}
\footnote{\href{https://github.com/MadryLab/robustness}{https://github.com/MadryLab/robustness}}.

We resize the input images to $256\times 265$ pixels and crop $224\times 224$ pixels from the center.
For our experiments on certification, we use a set of $500$ test images by choosing at most $1$ sample for each class.

\subsection{Choice of Adaptive BN hyper-parameters}
\label{sec_app_test_hyperparam}

BN adaptation technique is controlled by two hyper-parameters, i.e., the \textit{test batch-size} and \textit{momentum} ($\rho$) (see Equation \ref{eq_adaptStat}) to update the statistics of the batch-normalization layers.
Assuming that the test images are obtained independently from the same test distribution, we can efficiently compute the BN statistics from these images. 
The hyper-parameter $\rho \in [0,1]$ controls the tread-off between pre-computed training statistics and test statistics.
We can obtain a better estimation of the test distribution from a large test batch.
Hence, we can choose a higher value of $\rho$.

Here, we compare the top-1 test accuracy of AT models under Gaussian augmented noise with $\sigma= 0.5$ for different choices of $\rho$ and the batch size.
We skip the standard baseline models from our analysis and refer to the previous works  \citep{adaptBN_cp_nips_2020,adaptBN_cp_arxiv_2020} that analyzed the effects of these hyper-parameters for the standard baseline DNN classifiers.

\begin{table}[h]
\centering
\resizebox{12.0cm}{!} 
{%
\begin{tabular}{c|cclc|cc}
\cmidrule[1.2pt]{1-3} \cmidrule[1.2pt]{5-7}
\multicolumn{3}{c}{(a) {ImageNet}} & $\qquad$ &\multicolumn{3}{c}{(b) {CIFAR-10}} \\
\cmidrule[1.2pt]{1-3} \cmidrule[1.2pt]{5-7}
$\rho$ & Adv$_{\infty}$ & Adv$_2$ & \qquad &
$\rho$ & Adv$_{\infty}$ & Adv$_2$ 
\\ \cline{1-3} \cline{5-7}
0.0 (No adaptation) & 0.4{\tiny $\pm0.01$} & 0.9{\tiny $\pm0.01$} & &
0.0 (No adaptation) & 16.1{\tiny $\pm7.85$} & 21.5{\tiny $\pm7.79$}
\\
0.1 &	2.1{\tiny $\pm0.04$} & 7.7{\tiny $\pm0.09$} & &
0.1 &	45.1{\tiny $\pm0.49$} & 46.9{\tiny $\pm0.48$} 
\\
0.3 & 20.6{\tiny $\pm0.16$}& 36.6{\tiny $\pm0.09$} & &
0.3 & 59.2{\tiny $\pm0.42$}& 60.8{\tiny $\pm0.33$}
\\
0.5 & 41.1{\tiny $\pm0.09$}& 45.5{\tiny $\pm0.13$} & &
0.5 & 62.4{\tiny $\pm0.27$}& 64.4{\tiny $\pm0.6$} 
\\
0.7 & 43.5{\tiny $\pm0.14$}& 46.7{\tiny $\pm0.13$} & & 
0.7 & 62.8{\tiny $\pm0.52$}& 64.9{\tiny $\pm0.31$}
\\
0.9 & 44.2{\tiny $\pm0.12$}& 46.8{\tiny $\pm0.13$} & &
0.9 & 62.8{\tiny $\pm0.71$}& 64.9{\tiny $\pm0.31$}
\\
1.0 (Full adaptation) & 44.8{\tiny $\pm0.13$}& 47.2{\tiny $\pm0.14$} & &
1.0 (Full adaptation) & 62.4{\tiny $\pm0.64$}& 64.9{\tiny $\pm0.73$} 
\\ \cmidrule[1.2pt]{1-3} \cmidrule[1.2pt]{5-7}
\end{tabular}%
}
\vspace{-0.5em}
\caption{ Top-1 accuracy using fixed test batch-size $= 512$ for AT models under Gaussian augmented noise with $\sigma=0.5$ for different choices of momentum, $\rho$ during inference.
We randomly shuffle the test images to report ($mean+2\times sd$) of $5$ different runs.}
\label{table_app_rho}
\end{table}

\paragraph{Momentum ($\rho$). }
We first investigate the effect of momentum ($\rho$) as we choose a large batch size of $512$.
In Table \ref{table_app_rho}, we present the performance of AT models for different values of $\rho$.
Recall that, $\rho=1$ denotes \textit{full adaptation} (Equation \ref{eq_adaptStat}). 
Here, we completely ignore the training statistics and recompute the BN statistics using the test batches.
In contrast, $\rho=0$ represents \textit{no adaptation}, i.e., the same as the standard `deterministic' inference setup. 
In this case, we use the previously computed BN statistics obtained during training.

We observe that for {ImageNet} (Table \ref{table_app_rho} [Left]) the performance started converging at $\rho=0.7$.
For {CIFAR-10} (Table \ref{table_app_rho} [Right]), the convergence started even earlier at $\rho=0.5$.

\begin{table}[h]
\centering
\resizebox{10.0cm}{!} 
{%
\begin{tabular}{c|cclc|cc}
\cmidrule[1.2pt]{1-3} \cmidrule[1.2pt]{5-7}
\multicolumn{3}{c}{(a) {ImageNet}} & $\qquad$ &\multicolumn{3}{c}{(b) {CIFAR-10}} \\
\cmidrule[1.2pt]{1-3} \cmidrule[1.2pt]{5-7}
Batch Size & Adv$_{\infty}$ & Adv$_2$ & &
Batch Size & Adv$_{\infty}$ & Adv$_2$ 
\\ \cline{1-3} \cline{5-7}
w/o BN adapt & 0.4{\tiny $\pm0.01$} & 0.9{\tiny $\pm0.01$} & \qquad & 
w/o BN adapt & 16.1{\tiny $\pm7.85$} & 21.5{\tiny $\pm7.79$}
\\
8 &	11.5{\tiny $\pm0.22$}& 9.1{\tiny $\pm0.15$} & &
8 &	57.2{\tiny $\pm1.23$} & 59.5{\tiny $\pm0.38$}
\\
16 & 28.1{\tiny $\pm0.22$}& 26.7{\tiny $\pm0.14$} & &
16 & 60.2{\tiny $\pm0.79$} & 62.3{\tiny $\pm0.87$}
\\
32 & 37.1{\tiny $\pm0.24$}& 37.6{\tiny $\pm0.2$} & &
32 & 61.5{\tiny $\pm0.46$} & 63.6{\tiny $\pm0.55$}
\\
64 & 41.4{\tiny $\pm0.26$}& 42.9{\tiny $\pm0.12$} & & 
64 & 62.3{\tiny $\pm0.5$} & 64.0{\tiny $\pm0.38$}
\\
128 & 43.3{\tiny $\pm0.15$}& 45.4{\tiny $\pm0.13$} & &
128 & 62.7{\tiny $\pm0.68$} & 64.4{\tiny $\pm0.53$}
\\
256 & 44.4{\tiny $\pm0.21$}& 46.7{\tiny $\pm0.07$} & &
256 & 62.7{\tiny $\pm0.68$} & 64.9{\tiny $\pm0.48$}
\\
512 & 44.8{\tiny $\pm0.13$}& 47.2{\tiny $\pm0.14$} & &
512 & 62.4{\tiny $\pm0.64$}& 64.9{\tiny $\pm0.73$} 
\\ \cline{1-3} \cline{5-7}
\end{tabular}%
}
\caption{ Top-1 accuracy using fixed $\rho = 1$ for AT models under Gaussian augmented noise with $\sigma=0.5$ for different size of test batches during inference.
We randomly shuffle the test images to report ($mean+2\times s.d.$) of $5$ different runs.
}
\label{table_app_batchSize}
\end{table}

\paragraph{Batch Size. }
Next, we investigate the minimum size of the test batches to choose $\rho=1$ (i.e., full-adaptation).
In Table \ref{table_app_batchSize}, we fix $\rho=1$ and vary the test batch sizes as we evaluate these AT models.
We observe that the performance of these models started improving even when we are using the test batches of size $8$.
The performance further improves as we choose larger sizes of test batches. 
We can see that their performance started converging as we choose the test batches of size $64$ for {ImageNet}.
On the other hand, the convergence started much earlier for {CIFAR-10}.

\subsection{Performance against different corruptions}
\label{sec_app_corruption}
We mainly focus on $\ell_2$ certification using Gaussian noise in this paper.
However, we note that randomized smoothing techniques have been also applied to provide certifications for other perturbation types as well (e.g., random uniform noise for $\ell_1$ norm \citep{anyLpCertify_arxiv_2020}).
Consequently, we can apply our proposed Algorithm \ref{algo_certify} to adapt an AT model for any given perturbation types without any additional training for different applications.

Further, \cite{commonPerturbation_iclr_2019} recently introduced ImageNet-C and CIFAR10-C datasets by \textit{algorithmically generated random corruptions} from \textit{noise}, \textit{blur}, \textit{weather}, and \textit{digital} categories with $5$ different severity levels for each corruption.
Several recent works demonstrated that adaptive BN techniques can significantly improve the performance of any classifier (including AT models) against different random corruptions.
Further, -- also demonstrated the effectiveness of AT models even without applying any adaptation.
Hence, our proposed certification framework for AT models is a step forward towards further improving the reliability of sensitive real-world applications.

\end{document}